%% file: main.tex
\documentclass[10pt,twocolumn,letterpaper]{article}

\usepackage[pagenumbers]{cvpr} % To force page numbers, e.g. for an arXiv version
\usepackage{algorithm}
\usepackage{algorithmic}
\usepackage{multirow}
\usepackage{multicol}
\usepackage{makecell}
\usepackage{indentfirst} 

\input{preamble}

\definecolor{cvprblue}{rgb}{0.21,0.49,0.74}
\usepackage[pagebackref,breaklinks,colorlinks,citecolor=cvprblue]{hyperref}

\def\paperID{1891} % *** Enter the Paper ID here
\def\confName{CVPR}
\def\confYear{2024}

\title{RealNet: A Feature Selection Network with Realistic Synthetic Anomaly for Anomaly Detection}

\author{ Ximiao Zhang\textsuperscript{1} \hspace{0.8cm} Min Xu\textsuperscript{1}\thanks{Corresponding author.} \hspace{0.8cm} Xiuzhuang Zhou\textsuperscript{2}\\
\textsuperscript{1}College of Information and Engineering, Capital Normal University \\
\textsuperscript{2}School of Artificial Intelligence, Beijing University of Posts and Telecommunications\\
{\tt\small \{2211002048,\:xumin\}@cnu.edu.cn\textsuperscript{1}, xiuzhuang.zhou@bupt.edu.cn\textsuperscript{2}}}

\begin{document}
\maketitle

\begin{abstract}
  Self-supervised feature reconstruction methods have shown promising advances in industrial image anomaly detection and localization. Despite this progress, these methods still face challenges in synthesizing realistic and diverse anomaly samples, as well as addressing the feature redundancy and pre-training bias of pre-trained feature. In this work, we introduce RealNet, a feature reconstruction network with realistic synthetic anomaly and adaptive feature selection. It is incorporated with three key innovations: First, we propose Strength-controllable Diffusion Anomaly Synthesis (SDAS), a diffusion process-based synthesis strategy capable of generating samples with varying anomaly strengths that mimic the distribution of real anomalous samples. Second, we develop Anomaly-aware Features Selection (AFS), a method for selecting representative and discriminative pre-trained feature subsets to improve anomaly detection performance while controlling computational costs. Third, we introduce Reconstruction Residuals Selection (RRS), a strategy that adaptively selects discriminative residuals for comprehensive identification of anomalous regions across multiple levels of granularity. We assess RealNet on four benchmark datasets, and our results demonstrate significant improvements in both Image AUROC and Pixel AUROC compared to the current state-of-the-art methods. The code, data, and models are available at \url{https://github.com/cnulab/RealNet}.
\end{abstract}

\section{Introduction}

\begin{figure}[t]
  \centering
   \includegraphics[width=0.95\linewidth]{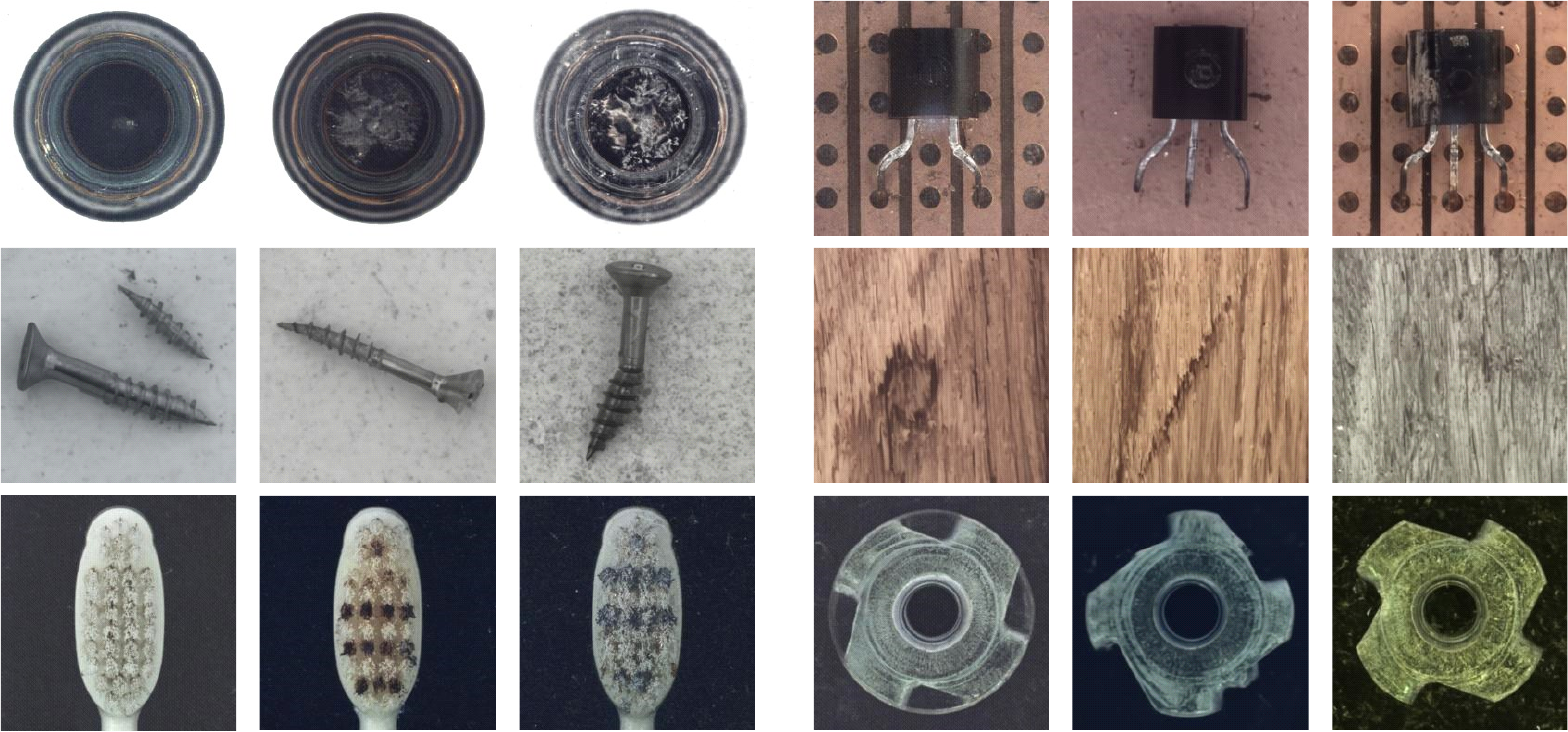}
   \caption{SDAS generates anomaly images using only normal images. The example images are sourced from the MVTec-AD dataset \cite{bergmann2019mvtec}.}
   \label{fig:fig1}
\end{figure}

Image anomaly detection is a critical task in industrial production, with wide-ranging applications in quality control and safety monitoring. While self-supervised methods \cite{zavrtanik2021draem,li2021cutpaste,schluter2022natural,zhang2023destseg,zhang2023prototypical} have gained attention for training models using synthetic anomalies, they still face challenges in synthesizing realistic and diverse anomaly images, especially generating complex structural anomalies and unseen anomaly categories. Due to the lack of available anomaly images and prior knowledge about anomaly categories, existing methods rely on carefully crafted data augmentation strategies \cite{li2021cutpaste,schluter2022natural} or external data \cite{zavrtanik2021draem} for anomaly synthesis, leading to significant distribution discrepancy between synthetic anomalies and real anomalies, thereby limiting the generalization ability of anomaly detection models to real-world applications. To address these issues, we introduce Strength-controllable Diffusion Anomaly Synthesis (SDAS), a novel synthesis strategy that generates diverse samples more closely aligned with natural distributions, and offers flexibility in controlling anomaly strength. SDAS employs DDPM \cite{ho2020denoising} to model the distribution of normal samples and introduces perturbation terms during the sampling process to generate samples in low probability density regions. These samples simulate various natural anomaly patterns, such as aging, structural changes, abnormal textures, and color changes, as shown in \cref{fig:fig1}. 

Parallel to this, feature reconstruction-based anomaly detection \cite{yang2020dfr,zavrtanik2022dsr,you2022unified,deng2022anomaly,zhang2023destseg} is another promising research direction, which reconstructs the features of anomalous images as those of normal images and conducts anomaly detection and localization by reconstruction residuals. They have attracted considerable attention due to the simple paradigm. However, due to the high computational demands of feature reconstruction and the lack of effective feature selection strategies, existing methods either employ small-scale pre-trained CNNs \cite{yang2020dfr,zavrtanik2022dsr,you2022unified} for anomaly detection or handpick layer-specific features from pre-trained network \cite{deng2022anomaly,zhang2023destseg} for reconstruction. The latest work \cite{heckler2023exploring} highlights the importance of feature selection, indicating that existing anomaly detection methods \cite{roth2022towards, yu2021fastflow} are sensitive to feature selection. The optimal pre-trained feature subset for anomaly detection varies across different categories. Therefore, devising a unified feature selection approach has become a pressing need for advancing anomaly detection. In this paper, we propose RealNet, a feature reconstruction framework that incorporates Anomaly-aware Features Selection (AFS) and Reconstruction Residuals Selection (RRS). RealNet fully exploits the discriminative capabilities of large-scale pre-trained CNNs while reducing feature redundancy and pre-training bias, enhancing anomaly detection performance while effectively controlling computational demands. For different categories, RealNet selects different pre-trained feature subsets for anomaly detection, ensuring optimal anomaly detection performance while flexibly controlling the model size. Furthermore, RealNet effectively reduces missed detections by adaptively discarding reconstruction residuals lacking anomalous information, and significantly improves the recall of anomalous regions. In summary, our contributions are fourfold:

\begin{itemize}
\item We propose RealNet, a feature reconstruction network that effectively leverages multi-scale pre-trained features for anomaly detection by adaptively selecting pre-trained features and reconstruction residuals. RealNet achieves state-of-the-art performance while addressing the computational cost limitations suffered by previous methods.
\item We introduce Strength-controllable Diffusion Anomaly Synthesis (SDAS), a novel anomaly synthesis strategy that generates realistic and diverse anomalous samples closely aligned with natural distributions.
\item We evaluate RealNet on four datasets (MVTec-AD \cite{bergmann2019mvtec}, MPDD \cite{jezek2021deep}, BTAD \cite{mishra2021vt}, and VisA \cite{zou2022spot}), surpassing existing state-of-the-art methods using the same set of network architectures and hyperparameters across datasets. 
\item We provide the Synthetic Industrial Anomaly Dataset (SIA). SIA is generated by SDAS and consists of a total of 360,000 anomalous images from 36 categories of industrial products. SIA can be conveniently utilized for anomaly synthesis to facilitate self-supervised anomaly detection methods.
\end{itemize}

\section{Related work}

% Unsupervised anomaly detection and localization approaches can be roughly classified into four main categories: 

Unsupervised anomaly detection and localization approaches use only normal images for model training, without any anomalous data. These methods can be roughly classified into four main categories: reconstruction-based methods \cite{akccay2019skip,baur2019deep}, self-supervised learning-based methods \cite{li2021cutpaste,zavrtanik2021draem}, deep feature embedding-based methods \cite{defard2021padim,roth2022towards}, and one-class classification-based methods \cite{yi2020patch,liznerski2021explainable}. In this paper, we focus on the reconstruction-based and self-supervised learning-based methods, which are of particular relevance to our proposed RealNet framework.

\textbf{Reconstruction-based methods} follow a relatively consistent paradigm, which entails training a reconstruction model on normal images. The inability to effectively reconstruct anomalous regions in input images facilitates anomaly detection and localization through comparison of the original and reconstructed images. In this context, a variety of reconstruction techniques are explored, such as Autoencoder \cite{baur2019deep,youkachen2019defect}, GAN \cite{akccay2019skip,schlegl2017unsupervised}, Transformer \cite{mishra2021vt,pirnay2022inpainting}, and Diffusion model \cite{wyatt2022anoddpm,Zhang2023ICCV,Lu2023ICCV}. However, managing the reconstruction capability of the network remains challenging. In cases of complex image structures or textures, the network may produce a simplistic copy instead of selective reconstruction. Furthermore, inherent stylistic discrepancies between original and reconstructed images can lead to false positives or undetected anomalies.

Recent studies, as exemplified by \cite{yang2020dfr,you2022unified,zavrtanik2022dsr,deng2022anomaly}, have been primarily focused on anomaly detection through the reconstruction of pre-trained image features. In contrast to image-level reconstruction, multi-scale features pre-trained on ImageNet \cite{deng2009imagenet} demonstrate enhanced discriminative abilities to detect anomalies across a wide range of scales and diverse image patterns. However, due to the inherent feature redundancy in high-dimensional features and the pre-training bias introduced by classification tasks, the anomaly detection capability of large-scale pre-trained networks has not been fully utilized. Recent studies \cite{yang2020dfr,zavrtanik2022dsr,you2022unified} use small-scale pre-trained networks to ensure controllable reconstruction costs, and other works \cite{zhang2023destseg,tien2023revisiting,roth2022towards} manually select partial layer features from pre-trained networks for anomaly detection. However, the optimal feature subset for anomaly detection varies across different categories \cite{heckler2023exploring}, thus, these manually selecting methods often prove to be dataset-specific and suboptimal, resulting in a significant performance drop. Different from previous solutions, our RealNet presents a novel combination of efficient feature selection strategies and an optimized reconstruction process, effectively enhancing anomaly detection performance while maintaining computational efficiency.

\begin{figure*}[t]
  \centering
   \includegraphics[width=\linewidth]{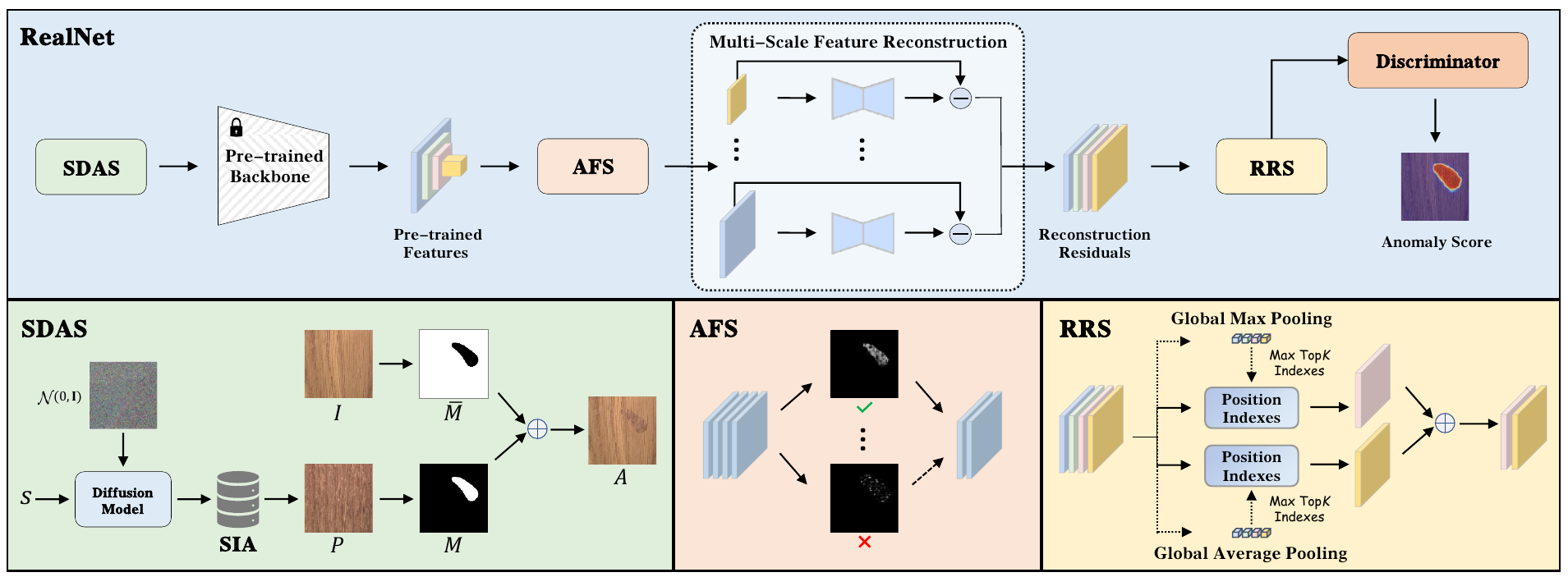}
   \caption{The pipeline of our RealNet consists of three core components: Strength-controllable Diffusion Anomaly Synthesis (SDAS), Anomaly-aware Features Selection (AFS), and Reconstruction Residuals Selection (RRS). 1) SDAS enables the synthesis of diverse, near-natural distribution anomalous images. 2) AFS refines features extracted by large-scale pre-trained CNN for dimensionality reduction. Refined features are reconstructed into corresponding normal image features by a set of reconstruction networks. 3) RRS selects reconstruction residuals most likely to identify anomalies, which are then fed into a discriminator for anomaly detection and localization.}
   \label{fig:fig2}
\end{figure*}

\textbf{Self-supervised learning-based methods} aim to bypass the need for labels of anomalous images by setting a suitable proxy task. Notable works in this domain include CutPaste \cite{li2021cutpaste}, which generates anomalies by transplanting image patches from one location to another, albeit with suboptimal continuity in the anomalous regions. NSA \cite{schluter2022natural} uses Poisson image editing \cite{perez2003poisson} for seamless image pasting to synthesize more natural anomaly regions. DRAEM \cite{zavrtanik2021draem} leverages the texture dataset DTD \cite{cimpoi2014describing} to synthesize various texture anomalies and achieve advanced self-supervised anomaly detection performance, however, it falls short when faced with specific structural anomalies, such as partial missing or misplaced elements.

The performance of self-supervised anomaly detection methods hinges on how closely the proxy task aligns with the real anomaly detection task. Anomaly synthesis, as a fundamental study in anomaly detection, has not yet received widespread exploration. Recent work \cite{duan2023few} use StyleGAN2 \cite{karras2020analyzing} for image editing to generate anomalous images. However, the proposed method relies on real anomalous images and cannot generate unseen anomaly types. In contrast, SDAS operates in the probability space, free from constraints imposed by data augmentation rules or existing data, enabling effective control over anomaly strengths and the generation of realistic and diverse anomaly images using only normal images.

\section{Method}

In this section, we will introduce our proposed feature reconstruction framework, RealNet, which consists of three key components: Strength-controllable Diffusion Anomaly Synthesis (SDAS), Anomaly-aware Features Selection (AFS), and Reconstruction Residuals Selection (RRS). The pipeline of RealNet is illustrated in \cref{fig:fig2}.

\subsection{Strength-controllable Diffusion Anomaly \\Synthesis}

Denoising Diffusion Probabilistic Models (DDPM) \cite{ho2020denoising} employ a forward diffusion process to incrementally add noise $\mathcal N(0,\textbf{I})$ to the original data distribution $q(x_0)$. At time $t$, the conditional probability distribution of the noisy data $x_t$ is $q(x_t|x_{t-1})=\mathcal N(x_t;\sqrt{1-\beta_t}x_{t-1},\beta_t\textbf{I})$, where $\{\beta_t\}_{t=1}^T$ is a fixed variance schedule, and $\{x_t\}_{t=1}^T$ are the latent variables. The diffusion process is defined as a Markov chain, with joint probability distribution $q(x_{1:T}|x_0)=\prod_{t=1}^{T}q(x_t|x_{t-1})$. Following the sum rule of Gaussian random variables, the conditional probability distribution of $x_t$ at time $t$ is $q(x_t|x_0)=\mathcal N(x_t;\sqrt{\bar{\alpha}_t}x_0, (1-\bar{\alpha}_t)\textbf{I})$, where $\alpha_t=1-\beta_t$, and $\bar{\alpha}_t=\prod_{i=1}^{t}\alpha_i$.

The reverse process is described as another Markov chain, where the mean and variance of the reverse process are parameterized by $\theta$, \ie, $p_\theta(x_{t-1}|x_t)=\mathcal N(x_{t-1};\mu_\theta(x_t,t),\Sigma_\theta(x_t,t))$. There are various ways to model $\mu_\theta(x_t,t)$; typically, neural networks $\epsilon_\theta(x_t,t)$ are used to model the noise $\epsilon$ in the diffusion process, resulting in $\mu_\theta(x_t,t)=\frac{1}{\sqrt{\alpha_t}}(x_t-\frac{\beta_t}{\sqrt{1-\bar{\alpha}_t}}\epsilon_\theta(x_t,t))$. In the training phase, our goal is to minimize the variational upper bound of the negative log-likelihood, which leads to the simplified objective:
\begin{equation}
\label{eq:equ1}
\mathcal{L}_{simple}=\mathbb{E}_{t,x_0,\epsilon}[\|\epsilon-\epsilon_\theta(x_t,t)\|^2]
\end{equation}

\begin{figure*}[t]
  \centering
   \includegraphics[width=0.99\linewidth]{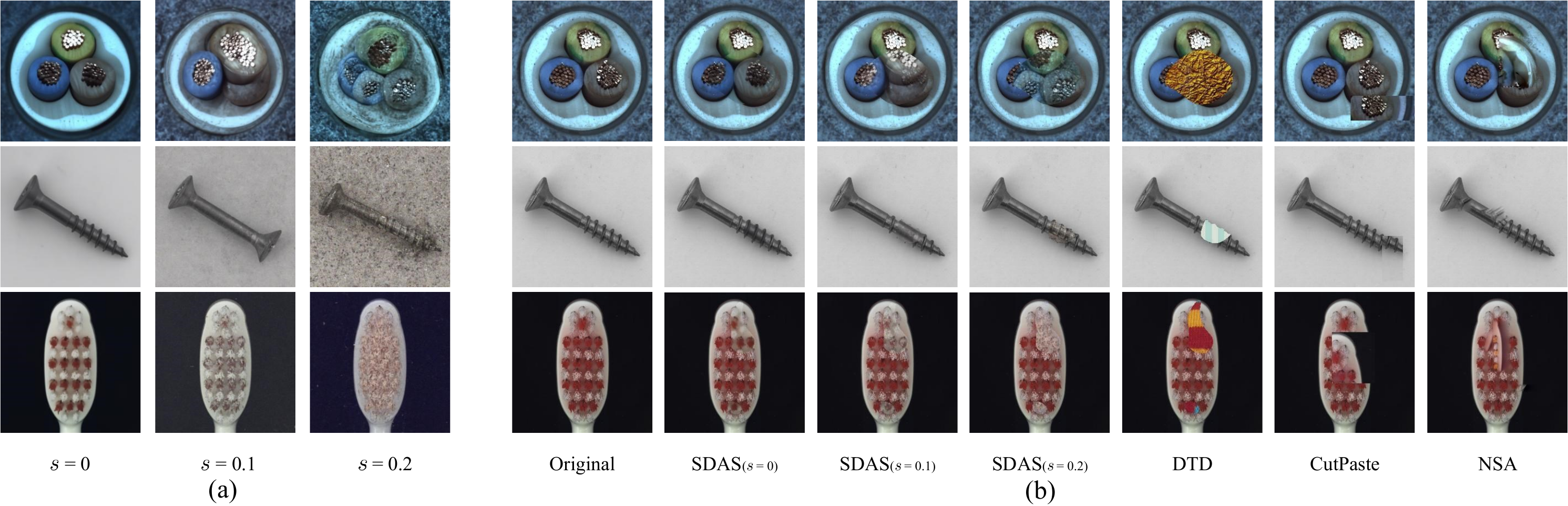}
   \caption{Anomaly image examples generated with different synthesis methods. (a) Examples generated using SDAS with different anomaly strengths $s$. (b) Examples featuring local anomaly regions generated by various anomaly synthesis methods.}
   \label{fig:fig3}
\end{figure*}

To generate realistic anomalous images, we first train a diffusion model to learn the distribution of normal images using \cref{eq:equ1}. In reverse diffusion process characterized by $p_\theta(x_{t-1}|x_{t})=\mathcal N(x_{t-1};\mu_\theta(x_t,t),\Sigma_\theta(x_t,t))$, $x_{t-1}$ is the normal image obtained at time $t-1$. Due to the anomalous images being located in low-density regions near the normal images, we introduce an additional perturbation $s\Sigma$ to sample anomalous images, yielding $p(x_{t-1}'|x_{t-1})=\mathcal N(x_{t-1}';x_{t-1},s\Sigma)$, where $\Sigma$ is the additional introduced variance, scalar $s$ controls the anomaly strength $(s \geq 0)$, and $x_{t-1}'$ is the anomalous image obtained at time $t-1$. To simplify the anomalous synthesis process, we set $\Sigma=\Sigma_\theta(x_t,t)$, by which the conditional probability distribution of anomalous images $x_{t-1}'$ can be written as follows:
\begin{equation}
\label{eq:equ2}
p_\theta(x_{t-1}'|x_{t})=\mathcal N(x_{t-1}';\mu_\theta(x_t,t),(1+s)\Sigma_\theta(x_t,t))
\end{equation}

\begin{algorithm}[b]
\caption{Strength-controllable Diffusion Anomaly Synthesis (SDAS)}
\begin{algorithmic}
\STATE \textbf{Input}: \text{diffusion model $(\mu_\theta(x_t,t),\Sigma_\theta(x_t,t))$}
\STATE \hspace{0.94cm} anomaly strength $s$
\STATE $x_T \sim \mathcal N(0,\textbf{I})$
\STATE \textbf{for all} $t$ \text{from T to 1} \textbf{do}
\STATE \hspace{0.4cm}$\mu,\Sigma \leftarrow \mu_\theta(x_t,t),\Sigma_\theta(x_t,t)$
\STATE \hspace{0.4cm}$x_{t-1} \sim \mathcal N(\mu,(1+s)\Sigma)$
\STATE \textbf{end for}
\STATE \textbf{return} $x_0$
\end{algorithmic}
\label{alg:algorithm1}
\end{algorithm}

To ensure that the generated anomalous images are close to the distribution of normal images, we set $s \rightarrow 0$, resulting in $x_{t-1}' \approx x_{t-1}$; then we use $x_{t-1}'$ for the next time step of the reverse diffusion process. The final form is $p_\theta(x_{t-1}'|x_{t}')=\mathcal N(x_{t-1}';\mu_\theta(x_t',t),(1+s)\Sigma_\theta(x_t',t))$. We term this process Strength-controllable Diffusion Anomaly Synthesis (SDAS), detailed in \cref{alg:algorithm1}. Specifically, SDAS will generate normal images if $s$ is set to 0.

To incorporate these anomalous images during training of anomaly detection model, we follow the approach presented in \cite{zavrtanik2021draem}, utilizing a Perlin noise generator \cite{perlin1985image} to capture various anomalous shapes and binarize them into an anomaly mask $M$. We denote the normal image as $I$, the anomalous image generated by SDAS as $P$, and the image with local anomalies synthesized by image blending as $A$:
\begin{equation}
\label{eq:equ3}
A=\overline{M} \odot I+(1-\delta)(M \odot I)+\delta(M \odot P)
\end{equation}
where $\overline{M}=1-M$, $\odot$ denotes the element-wise multiplication operation, and $\delta$ is the opacity in the image blending. To ensure that the generated anomalous regions are located in the foreground, we use an adaptive threshold-based binarization method for foreground segmentation, similar to methods used in \cite{yang2023memseg,yao2023explicit,schluter2022natural}. \cref{fig:fig3}a shows the images generated by SDAS under different anomaly strengths, while \cref{fig:fig3}b compares the images with local anomaly regions synthesized by different methods. The larger the value of $s$, the greater the distribution difference between the generated image and the normal image, and the more obvious the abnormal region obtained after image blending. When $s$ is very small, imperceptible abnormal regions can be synthesized. Compared with alternative synthesis methods, the anomalies generated by SDAS are more continuous and can have very realistic structural anomalies. 

\subsection{Anomaly-aware Features Selection}

In this section, we introduce the Anomaly-aware Features Selection (AFS) module within RealNet, a self-supervised method for pre-trained feature selection, reducing feature dimensionality and eliminating pre-training bias, as well as managing reconstruction costs. Firstly, we define a set of $N$ triplets $\{A_n,I_n,M_n\}_{n=1}^N$, where $A_n,I_n\in R^{h\times w\times 3}$ represent anomaly images synthesized by SDAS and original normal images, and $M_n\in R^{h\times w}$ represents the corresponding anomaly mask. We denote the pre-trained network as $\phi_k$, and $\phi_k(A_n)\in R^{h_k\times w_k\times c_k}$ represents the $k$th layer pre-trained feature extracted from $A_n$, where $c_k$ represents the number of channels. For the $i$th feature map, $\phi_{k,i}(A_n)\in R^{h_k\times w_k}$, AFS selects $m_k$ feature maps for reconstruction ($m_k \leq c_k$). Specifically, the feature maps indexed by $k$ are from ResNet-like architectures, such as ResNet50 \cite{he2016deep} or WideResNet50 \cite{zagoruyko2016wide}, where $k \in \{1,2,3,4\}$ represent the last layer outputs of blocks with different spatial resolutions.

For the $k$th layer pre-trained features, we define the following AFS loss for evaluation of the $i$th feature map:
\begin{equation}
\label{eq:equ4}
\mathcal{L}_{AFS}(\phi_{k,i})=\frac{1}{N}\sum_{n=1}^N\|F([\phi_{k,i}(A_n)-\phi_{k,i}(I_n)]^2)-M_n\|_2^2
\end{equation}
where $F(\cdot)$ is a function that performs normalization operation and aligns the resolution of $[\phi_{k,i}(A_n)-\phi_{k,i}(I_n)]^2$ to $M_n$. Given the feature reconstruction process for anomalous images, we train a reconstruction network to infer $\phi_{k,i}(I_n)$ based on $\phi_{k,i}(A_n)$, which enables the detection and localization of anomalies through $[\phi_{k,i}(A_n)-\phi_{k,i}(I_n)]^2$. Ideally, $[\phi_{k,i}(A_n)-\phi_{k,i}(I_n)]^2$ should closely approximate $M_n$. The $\mathcal{L}_{AFS}(\phi_{k,i})$ represents the capability of $\phi_{k,i}$ in identifying anomalous regions. Due to the unavailability of real anomalous samples, we employ synthetic anomalies for feature selection. For the $k$th layer of pre-trained features, AFS selects $m_k$ feature maps with the smallest $\mathcal{L}_{AFS}$ for reconstruction. We denote the AFS as $\varphi_k(\cdot)$, and $\varphi_k(A_n) \in R^{h_k\times w_k\times m_k}$, where $m_k \leq c_k$. We perform AFS on each layer of pre-trained features separately, and finally obtain selected multi-scale features $\{\varphi_1(A_n),...,\varphi_K(A_n)\}$. In this process, each layer's feature dimension $\{m_1,...,m_K\}$ serves as a set of hyperparameters. Specifically, for RealNet, AFS operation is performed only once on the pre-trained features of each layer, and the index of the selected feature maps is cached for subsequent training and inference.

AFS adaptively selects a subset of features from all available layers for anomaly detection, offering the following advantages compared to conventional methods \cite{zhang2023destseg,tien2023revisiting,roth2022towards} that select all features from partial layers: 1) AFS reduces feature redundancy within layers and mitigates pre-training bias, enhancing both feature representativeness and discriminability to improve anomaly detection performance. 2) AFS broadens the receptive field to enhance multi-scale anomaly detection capabilities. 3) AFS distinguishes the dimensions of pre-trained features from those employed for anomaly detection, ensuring efficient control over computational costs and flexible customization of the model size.

In RealNet, a set of reconstruction networks $\{G_1,...,G_K\}$ are designed to reconstruct the selected synthetic anomalous features $\{\varphi_1(A_n),...,\varphi_K(A_n)\}$ into the original image features $\{\varphi_1(I_n),...,\varphi_K(I_n)\}$ at various resolutions. The loss function $\mathcal{L}_{recon}$ is defined as:

\begin{equation}
\label{eq:equ5}
\mathcal{L}_{recon}(A,I)=\frac{1}{N}\sum_{n=1}^N\sum_{k=1}^K\|G_k(\varphi_k(A_n))-\varphi_k(I_n)\|_2^2
\end{equation}

During the reconstruction process, we intentionally forgo aligning multi-scale features \cite{yang2020dfr,you2022unified,roth2022towards} to preserve optimal performance. This choice is motivated by the potential drawbacks associated with aligning low-resolution features through down-sampling, which could compromise the network's detection resolution and increase the risk of misidentifying anomalies. On the other hand, aligning high-resolution features using up-sampling may result in unnecessary feature redundancy, leading to elevated reconstruction costs. A detailed discussion on the reconstruction network architectures can be found in \cref{sec: supC}.

\begin{table*}[t]
  \renewcommand\arraystretch{1.0}
  \centering
  \caption{Comparison of SIA with alternative anomaly synthesis approaches on the MVTec-AD dataset \cite{bergmann2019mvtec}, employing Image AUROC (\%), Pixel AUROC (\%), and PRO (\%) as evaluation metrics. }
    \resizebox{0.91\linewidth}{!}{
    \tiny
    \begin{tabular}{c|c|c|c|c|c}
    \bottomrule
    \multicolumn{2}{c|}{Category} & SIA   & DTD \cite{cimpoi2014describing}   & NSA \cite{schluter2022natural}   & CutPaste \cite{li2021cutpaste} \\
    \hline
    \multicolumn{1}{c|}{\multirow{6}{*}{Texture}} & \multicolumn{1}{c|}{Carpet} & (99.84, 99.19, 96.41) & (\textbf{100.0}, \textbf{99.27}, \textbf{96.96}) & (99.80, 98.60, 88.77) & (99.24, 98.42, 93.85) \\
\cline{2-6}    \multicolumn{1}{c|}{} & \multicolumn{1}{c|}{Grid} & (\textbf{100.0}, 99.51, \textbf{97.28}) & (\textbf{100.0}, \textbf{99.57}, 97.14) & (\textbf{100.0}, 99.32, 91.31) & (\textbf{100.0}, 99.18, 92.53) \\
\cline{2-6}    \multicolumn{1}{c|}{} & \multicolumn{1}{c|}{Leather} & (\textbf{100.0}, 99.76, 96.22) & (\textbf{100.0}, \textbf{99.77}, 96.41) & (\textbf{100.0}, 99.24, \textbf{96.85}) & (\textbf{100.0}, 99.41, 92.12) \\
\cline{2-6}    \multicolumn{1}{c|}{} & \multicolumn{1}{c|}{Tile} & (99.96,\textbf{99.44}, \textbf{97.70}) & (\textbf{100.0}, 99.35, 95.27) & (\textbf{100.0}, 97.40, 86.45) & (99.86, 97.63, 84.39) \\
\cline{2-6}    \multicolumn{1}{c|}{} & \multicolumn{1}{c|}{Wood} & (99.21, 98.22, 90.54) & (\textbf{99.65}, \textbf{98.28}, \textbf{91.23}) & (97.63, 93.30, 87.20) & (98.95, 95.29, 81.47) \\
\cline{2-6}    \multicolumn{1}{c|}{} & \multicolumn{1}{c|}{\textbf{AVG}} & (99.80, 99.22, \textbf{95.63}) & (\textbf{99.93}, \textbf{99.25},95.40) & (99.49, 97.57, 90.11) & (99.61, 97.99, 88.87) \\
    \hline
    \multicolumn{1}{c|}{\multirow{11}{*}{Object}} & \multicolumn{1}{c|}{Bottle} & (\textbf{100.0}, 99.30, \textbf{95.62}) & (\textbf{100.0}, 99.35, 95.57) & (\textbf{100.0}, \textbf{99.37}, 93.49) & (\textbf{100.0}, 99.14, 91.41) \\
\cline{2-6}    \multicolumn{1}{c|}{} & \multicolumn{1}{c|}{Cable} & (99.19, \textbf{98.10}, \textbf{93.38}) & (98.95, 97.84, 90.36) & (\textbf{99.33}, 97.62, 93.26) & (96.35, 96.23, 86.05) \\
\cline{2-6}    \multicolumn{1}{c|}{} & \multicolumn{1}{c|}{Capsule} & (\textbf{99.56}, \textbf{99.32}, 84.48) & (99.32, 99.19, 82.28) & (99.04, 99.27, \textbf{85.77}) & (98.48, 99.10, 79.55) \\
\cline{2-6}    \multicolumn{1}{c|}{} & \multicolumn{1}{c|}{hazelnut} & (\textbf{100.0}, \textbf{99.68}, 93.14) & (\textbf{100.0}, 99.46, 93.46) & (\textbf{100.0}, 99.25, \textbf{94.41}) & (\textbf{100.0}, 99.03, 91.51) \\
\cline{2-6}    \multicolumn{1}{c|}{} & \multicolumn{1}{c|}{Metal Nut} & (99.76, 98.58, 94.39) & (99.90, 98.58, \textbf{96.49}) & (\textbf{100.0}, \textbf{99.11}, 93.27) & (99.90, 98.03, 89.69) \\
\cline{2-6}    \multicolumn{1}{c|}{} & \multicolumn{1}{c|}{Pill} & (\textbf{99.13}, \textbf{99.02}, 91.04) & (98.36, 98.88, 84.44) & (97.19, 98.28, \textbf{95.15}) & (97.22, 98.96, 86.48) \\
\cline{2-6}    \multicolumn{1}{c|}{} & \multicolumn{1}{c|}{Screw} & (\textbf{98.83}, 99.45, 87.90) & (97.72, 99.36, 85.22) & (98.79, \textbf{99.62}, \textbf{93.74}) & (92.74, 98.53, 79.63) \\
\cline{2-6}    \multicolumn{1}{c|}{} & \multicolumn{1}{c|}{Toothbrush} & (99.44, 98.71, \textbf{91.57}) & (99.44, 98.69, 90.87) & (\textbf{100.0}, \textbf{99.18}, 89.20) & (99.17, 98.85, 78.48) \\
\cline{2-6}    \multicolumn{1}{c|}{} & \multicolumn{1}{c|}{Transistor} & (\textbf{100.0}, \textbf{98.00}, \textbf{92.92}) & (99.71, 97.15, 86.56) & (98.54, 95.67, 79.09) & (99.38, 96.32, 76.52) \\
\cline{2-6}    \multicolumn{1}{c|}{} & \multicolumn{1}{c|}{zipper} & (99.82, \textbf{99.17}, \textbf{93.43}) & (99.68, 99.02, 88.77) & (\textbf{99.90}, 98.91, 93.05) & (99.61, 98.03, 92.26) \\
\cline{2-6}    \multicolumn{1}{c|}{} & \multicolumn{1}{c|}{\textbf{AVG}} & (\textbf{99.57}, \textbf{98.93}, \textbf{91.79}) & (99.31, 98.75, 89.40) & (99.28, 98.63, 91.04) & (98.29, 98.22, 85.16) \\
    \hline
    \multicolumn{2}{c|}{\textbf{AVG}} & (\textbf{99.65}, \textbf{99.03}, \textbf{93.07}) & (99.52, 98.92, 91.40) & (99.35, 98.28, 90.73) & (98.73, 98.14, 86.40) \\
    \toprule
    \end{tabular}
    }
  \label{tab:table1}
\end{table*}

\begin{table*}[t]
  \renewcommand\arraystretch{1.1}
  \centering
  \caption{Comparison of RealNet with alternative anomaly detection methods on the MVTec-AD dataset \cite{bergmann2019mvtec}.}
   \resizebox{0.91\linewidth}{!}{
    \begin{tabular}{c|ccc|cccccc|c}
    \bottomrule
    \multicolumn{1}{c|}{Metric} & \multicolumn{1}{c}{\textit{PatchCore} \cite{roth2022towards}} & \multicolumn{1}{c}{\textit{SimpleNet} \cite{liu2023simplenet}} & \multicolumn{1}{c|}{\textit{FastFlow} \cite{yu2021fastflow}} & \multicolumn{1}{c}{DRAEM+SSPCAB \cite{ristea2022self}} & DSR \cite{zavrtanik2022dsr}   & \multicolumn{1}{c}{UniAD \cite{you2022unified}} & \multicolumn{1}{c}{RD++ \cite{tien2023revisiting}} & \multicolumn{1}{c}{DeSTSeg \cite{zhang2023destseg}} & \multicolumn{1}{c|}{DiffAD \cite{Zhang2023ICCV}} & \multicolumn{1}{c}{RealNet} \\
    \hline
    Image AUROC & 99.1 &\textbf{99.6} & 99.3  & 98.9  & \multicolumn{1}{c}{98.2} & 96.6  & 99.4  & 98.6  & 98.7 & \textbf{99.6} \\
    Pixel AUROC & 98.1 &98.1 & 98.1  & 97.2  & -     & 96.6  & 98.3  & 97.9  & 98.3 & \textbf{99.0} \\
    \toprule
    \end{tabular}
    }
  \label{tab:table2}
\end{table*}

\subsection{Reconstruction Residuals Selection}

In this section, we present the Reconstruction Residuals Selection (RRS) module. Reconstruction residuals are denoted as $\{E_1(A_n),...,E_K(A_n)\}$, where $E_k(A_n)=[\varphi_k(A_n)-G_k(\varphi_k(A_n))]^2$. To obtain the global reconstruction residual $E(A_n)\in R^{h'\times w'\times m'}$, we up-sample the low-resolution reconstruction residuals and concatenate them channel-wise, where $m'=\sum_{k=1}^Km_k$, $h'=max(h_1,...,h_K)$, and $w'=max(w_1,...,w_K)$.

The reconstruction residuals in $E(A_n)$ is obtained from the pre-trained features of reconstructing corresponding layer, and the features of the same resolution only have good ability to capture anomalies within a certain range. For instance, subtle low-level texture anomalies can be effectively captured exclusively by reconstruction residuals derived from low-level features. Therefore, RRS selects only a subset of reconstruction residuals that contain the most anomalous information for the anomaly score generation, to achieve the highest possible recall of anomalous regions.

Firstly, RRS performs GlobalMaxPooling (GMP) and GlobalAveragePooling (GAP) on $E(A_n)$ to obtain $E_{GMP}(A_n), E_{GAP}(A_n) \in R^{m'}$ respectively. The $r$ largest elements in $E_{GMP}(A_n)$ and $E_{GAP}(A_n)$ are then used to index the positions of $E(A_n)$ and obtain $E_{max}(A_n,r), E_{avg}(A_n,r) \in R^{h' \times w' \times r}$, which respectively represent the Top\textit{K} reconstruction residuals with the highest maximum and average values. To avoid missed detections caused by inadequate resolution, reconstruction residuals with insufficient anomalous information are discarded in RRS.

As GMP and GAP respectively represent local and global properties spatially, $E_{max}$ is more effective in capturing local anomalies in small areas, while $E_{avg}$ focuses on selecting anomalies with large spans. Combining $E_{max}$ and $E_{avg}$ together can enhance the RRS's ability to capture anomalies of various scales. We define the RRS operator as $E_{RRS}(A_n,r) \in R^{h'\times w' \times r}$. $E_{RRS}(A_n,r)$ concatenates $E_{max}(A_n,r/2)$ and $E_{avg}(A_n,r/2)$. Finally, we feed the $E_{RRS}(A_n,r)$ into a discriminator, which maps the reconstruction residual to the image-level resolution, obtaining the final anomaly scores. The maximum value in anomaly scores is used as the image-level anomaly score. We use cross entropy loss $\mathcal{L}_{seg}(A,M)$ to supervise the training of discriminator. The overall loss of RealNet is:
\begin{equation}
\label{eq:equ6}
\mathcal{L}(A,I,M)=\mathcal{L}_{recon}(A,I)+\mathcal{L}_{seg}(A,M)
\end{equation}

\subsection{Synthetic Industrial Anomaly Dataset}

To facilitate the reuse of generated anomaly images by SDAS, we constructed the Synthetic Industrial Anomaly Dataset (SIA). SIA comprises anomaly images for 36 categories from four industrial anomaly detection datasets, including MVTec-AD \cite{bergmann2019mvtec}, MPDD \cite{jezek2021deep}, BTAD \cite{mishra2021vt}, and VisA \cite{zou2022spot}. We generated 10,000 anomaly images with a resolution of $256 \times 256$ for each category, with anomaly strength $s$ uniformly sampled between 0.1 and 0.2. SIA can be conveniently used for synthesizing anomaly images through image blending, as described in \cref{eq:equ3}, and can serve as an effective alternative to the widely used DTD dataset \cite{cimpoi2014describing}.

\section{Experiment}

\subsection{Experimental setup}

\label{setup}

\textbf{Datasets.} We conduct extensive evaluations on four datasets, including MVTec-AD \cite{bergmann2019mvtec}, MPDD \cite{jezek2021deep}, BTAD \cite{mishra2021vt}, and VisA \cite{zou2022spot}. MVTec-AD \cite{bergmann2019mvtec} contains 5,354 images from 15 categories for industrial anomaly detection tasks, including 10 object categories and 5 texture categories. MPDD \cite{jezek2021deep} contains 1,346 images from 6 types of industrial metal products with varying lighting conditions, non-uniform backgrounds, and multiple products in each image. Furthermore, the placement orientation, shooting distance, and position of the products are also varied. BTAD \cite{mishra2021vt} contains images of 3 industrial products from the real world. VisA \cite{zou2022spot} is comprised of 9,621 normal images and 1,200 anomaly images from 12 categories. Certain categories demonstrate intricate structures, as exemplified by PCBs, while others consist of multiple objects that require detection, such as Capsules, thus rendering detection and localization a challenging task.

\textbf{Metrics.} To evaluate the performance of image-level anomaly detection, we use the Area Under the Receiver Operator Curve (AUROC) metric, as in previous works \cite{bergmann2019mvtec,jezek2021deep,mishra2021vt,zou2022spot}. For pixel-level anomaly location, we use Pixel AUROC and Per Region Overlap (PRO) \cite{bergmann2020uninformed}.

\begin{figure*}[t]
\centering
\begin{minipage}{0.48\textwidth}
		\centering
		\makeatletter\def\@captype{table}\makeatother
        \renewcommand\arraystretch{1.05}
        \caption{Comparison of SIA with DTD \cite{cimpoi2014describing} and CutPaste \cite{li2021cutpaste} on the MPDD dataset \cite{jezek2021deep}, employing Image AUROC (\%), Pixel AUROC (\%), and PRO (\%) as evaluation metrics.}
       \resizebox{\linewidth}{!}{
        \begin{tabular}{c|c|c|c}
            \bottomrule
            Category & SIA   & DTD \cite{cimpoi2014describing}   & CutPaste \cite{li2021cutpaste} \\
            \hline
            Bracket Black & (\textbf{94.95}, \textbf{99.27}, 87.10) & (89.49, 98.90, \textbf{88.57}) & (66.42, 96.67, 56.53) \\
            \hline
            Bracket Brown & (\textbf{96.83}, \textbf{97.81}, \textbf{94.36}) & (92.99, 97.35, 92.64) & (95.48, 97.54, 55.17) \\
            \hline
            Bracket White & (\textbf{88.78}, 97.44, \textbf{84.00}) & (86.67, \textbf{98.59}, 77.08) & (88.44, 96.51, 64.32) \\
            \hline
            Connector & (\textbf{100.0}, 97.46, \textbf{84.79}) & (99.05, 97.76, 65.91) & (99.05, \textbf{98.47}, 74.05) \\
            \hline
            Metal Plate & (\textbf{100.0}, 99.28, \textbf{94.44}) & (\textbf{100.0}, \textbf{99.35}, 93.78) & (99.95, 98.83, 92.69) \\
            \hline
            Tubes & (\textbf{97.51}, 97.94, 93.29) & (92.62, \textbf{99.01}, \textbf{96.49}) & (91.49, 98.09, 92.99) \\
            \hline
            \textbf{AVG}   & (\textbf{96.35}, 98.20, \textbf{89.66}) & (93.47, \textbf{98.49}, 85.75) & (90.14, 97.69, 72.63) \\
            \toprule
            \end{tabular}%
        }
        \label{tab:table3}
\end{minipage}\quad
\begin{minipage}{0.48\textwidth}
        \centering
        \makeatletter\def\@captype{figure}\makeatother
    \includegraphics[width=0.94\linewidth]{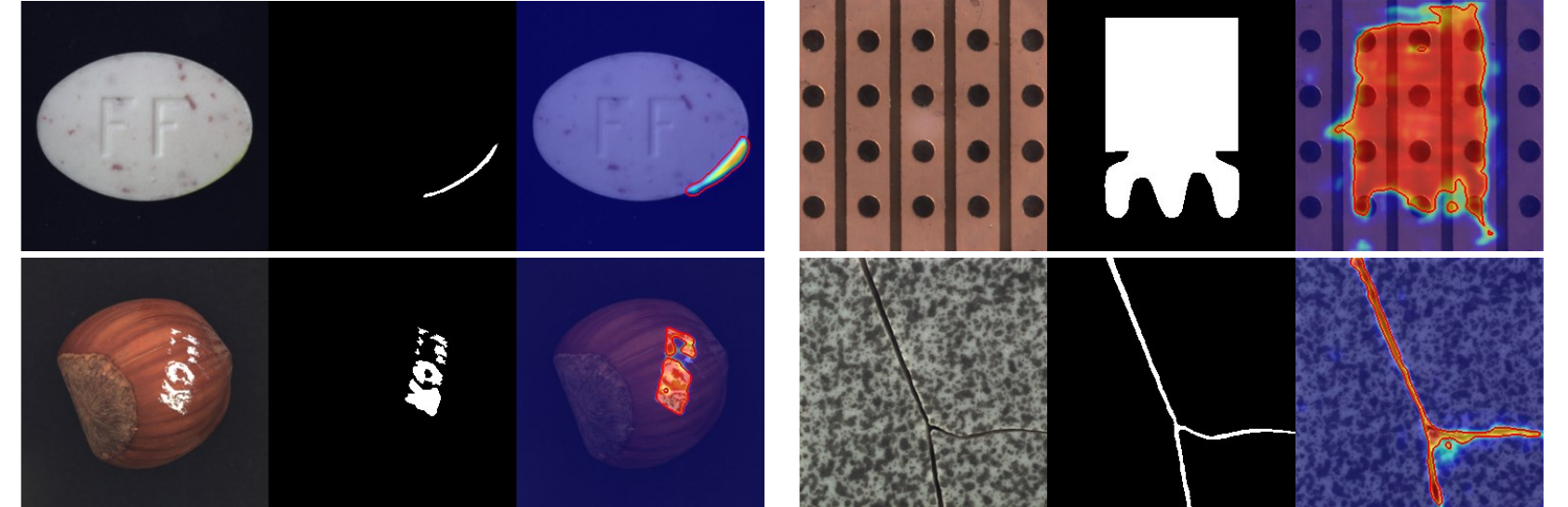}
    \caption{Qualitative results of RealNet on the MVTec-AD dataset \cite{bergmann2019mvtec}. Within each group, from left to right, are the anomaly image, ground-truth, and predicted anomaly score.}
    \label{fig:fig4}
\end{minipage}
\end{figure*}

\begin{table*}[h]
  \centering
  \renewcommand\arraystretch{1.0}
    \caption{Comparison of RealNet with alternative anomaly detection methods on the MPDD dataset \cite{jezek2021deep}.}
   \resizebox{0.9\linewidth}{!}{
       \tiny
    \begin{tabular}{c|cccc|cc|c}
    \bottomrule
    \multicolumn{1}{c|}{Metric} & \multicolumn{1}{c}{\textit{PatchCore} \cite{roth2022towards}} & \multicolumn{1}{c}{\textit{CFlow} \cite{gudovskiy2022cflow}} & \multicolumn{1}{c}{\textit{PaDiM} \cite{defard2021padim}} & \multicolumn{1}{c|}{\textit{SPADE} \cite{cohen2020sub}} & \multicolumn{1}{c}{DAGAN \cite{tang2020anomaly}} & \multicolumn{1}{c|}{Skip-GANomaly \cite{akccay2019skip}} & \multicolumn{1}{c}{RealNet} \\
    \hline
    Image AUROC & 82.1  & 86.1  & 74.8  & 77.1  & 72.5  & 64.8  & \textbf{96.3} \\
    Pixel AUROC & 95.7  & 97.7  & 96.7  & 95.9  & 83.3  & 82.2  & \textbf{98.2} \\
    \toprule
    \end{tabular}
  }
  \label{tab:table4}
\end{table*}

\textbf{Implementation details.} We evaluate RealNet on four datasets with consistent network architectures and hyperparameters, without specific tuning for individual categories. We use a WideResNet50 \cite{zagoruyko2016wide} pre-trained on ImageNet \cite{deng2009imagenet} as the backbone. In AFS, we set the dimension of pre-trained feature of each layer to $\{256,512,512,256\}$ for reconstruction. For RRS, $1/3$ of the reconstruction residuals are reserved to generate the final anomaly scores. For SDAS, we train the diffusion model following \cite{dhariwal2021diffusion} and use the SIA dataset for anomaly synthesis. Both SDAS and anomaly detection are performed at a resolution of $256\times256$ without center cropping, with a batch size of 16, and we use 64 batches of synthetic anomaly images for AFS. More details can be found in \cref{sec: supB}.

\subsection{Anomaly detection on MVTec-AD}

We train RealNet using SIA and alternative anomaly synthetic methods on the MVTec-AD dataset \cite{bergmann2019mvtec}, to evaluate the model's performance in anomaly detection and localization. These methods include: 1) \textbf{DTD} \cite{cimpoi2014describing}: This method utilizes the DTD dataset \cite{cimpoi2014describing} to blend images with generated anomalous textures, and the data augmentation strategy in \cite{zavrtanik2021draem} is employed during the blending process. 2) \textbf{NSA} \cite{schluter2022natural}: This method employs Poisson image editing \cite{perez2003poisson} to seamless image editing, following parameter setting in \cite{schluter2022natural}. 3) \textbf{CutPaste} \cite{li2021cutpaste}: This method involves random image cropping and pasting to synthesize anomaly regions. 

The experimental results are shown in \cref{tab:table1}. SDAS demonstrates flexibility in controlling the anomaly strength and generates synthetic anomalies with multiple anomaly patterns, especially for the object categories, where it achieves the best detection and localization performance. Compared to other methods, SDAS is not constrained by data augmentation rules or external data, enabling the synthesis of more natural and rich functional anomalies, as shown in \cref{fig:fig3}. RealNet trained using SIA achieves remarkable performance on the MVTec-AD dataset \cite{bergmann2019mvtec}, with an Image AUROC of 99.65\%, a Pixel AUROC of 99.03\%, and a PRO score of 93.07\%. \cref{fig:fig4} presents the qualitative anomaly localization results of RealNet on the MVTec-AD dataset \cite{bergmann2019mvtec}. The method exhibits remarkable pixel-level anomaly localization, proficiently identifying diverse anomaly patterns at various scales. Furthermore, RealNet can achieve a rapid inference speed of 31.93 FPS when using a single Nvidia GeForce RTX 3090, and it can perform inference using only 4GB of GPU memory. A detailed computational efficiency analysis can be found in \cref{sec: supC}.

We also compare RealNet with several state-of-the-art anomaly detection methods, and the results are shown in \cref{tab:table2}. Built on the same pre-trained network, RealNet outperforms the state-of-the-art alternatives, including Deep Feature Embedding-Based method (PatchCore \cite{roth2022towards} and SimpleNet \cite{liu2023simplenet}) and the NF-Based method (FastFlow \cite{yu2021fastflow}). When compared to previous reconstruction-based methods, RealNet achieves significant performance improvement.

\begin{table*}[h]
	\centering
    \label{tab:table5}
    \caption{Ablation studies of RealNet on the MVTec-AD dataset \cite{bergmann2019mvtec}.}
    \begin{subtable}{0.47\textwidth}
        \renewcommand\arraystretch{0.975}
        \caption{The impact of AFS and RRS on RealNet.}
        \label{tab:table5-a}
        \resizebox{\linewidth}{!}{
            \tiny
            \begin{tabular}{c|cc|c|c|c}
            \bottomrule
                  & AFS   & RRS   & \multicolumn{1}{c|}{Image AUROC} & \multicolumn{1}{c|}{Pixel AUROC} & \multicolumn{1}{c}{PRO} \\
            \hline
            1     & -     & -     & \multicolumn{1}{c|}{94.46 / 95.67} & \multicolumn{1}{c|}{93.38 / 95.84} & \multicolumn{1}{c}{79.81 / 82.26} \\
            2     & \checkmark     & -     & 96.86 & 96.32 & 84.13 \\
            3     & -     & \checkmark     & \multicolumn{1}{c|}{99.39 / 99.09} & \multicolumn{1}{c|}{98.66 / 98.22} & \multicolumn{1}{c}{92.01 / 88.38} \\
            4     & \checkmark     & \checkmark     & \textbf{99.65} & \textbf{99.03} & \textbf{93.07} \\
            \toprule
            \end{tabular}%
            }
	\end{subtable}\quad
	\begin{subtable}{0.50\textwidth}
        \renewcommand\arraystretch{1.02}
        \caption{The impact of anomaly strengths on RealNet.}
        \label{tab:table5-b}
        \resizebox{\linewidth}{!}{
        \tiny
    	 \begin{tabular}{c|c|c|c|c}
        \bottomrule
        \multicolumn{1}{c|}{Metric} & $s$ = 0     & $s$ = 0.1  & $s$ = 0.2   & \multicolumn{1}{c}{$s \in$ [0.1, 0.2]} \\
        \hline
        Image AUROC & 99.35 & \textbf{99.65} & 99.61 & \textbf{99.65} \\
        Pixel AUROC & 98.85 & 98.96 & 98.95 & \textbf{99.03} \\
        PRO   & 91.80  & 92.16 & 89.36 & \textbf{93.07} \\
        \toprule
        \end{tabular}%
        }
	\end{subtable}\quad
\end{table*}%

\subsection{Anomaly detection on MPDD}

We evaluate RealNet on MPDD dataset \cite{jezek2021deep} with SIA, DTD \cite{cimpoi2014describing}, and CutPaste \cite{li2021cutpaste}, and the results are shown in \cref{tab:table3}. Notably, RealNet trained with SIA achieves a significant improvement of 2.88\% in Image AUROC over DTD \cite{cimpoi2014describing}. \cref{tab:table4} shows the Image AUROC and the Pixel AUROC of RealNet and other methods on the MPDD dataset \cite{jezek2021deep}. RealNet achieves an Image AUROC of 96.3\%, surpassing the current best performance (CFlow \cite{gudovskiy2022cflow}) by 10.2\%, even without any dataset-specific tuning.

\begin{figure*}[t]
  \centering
  \includegraphics[width=0.919\textwidth]{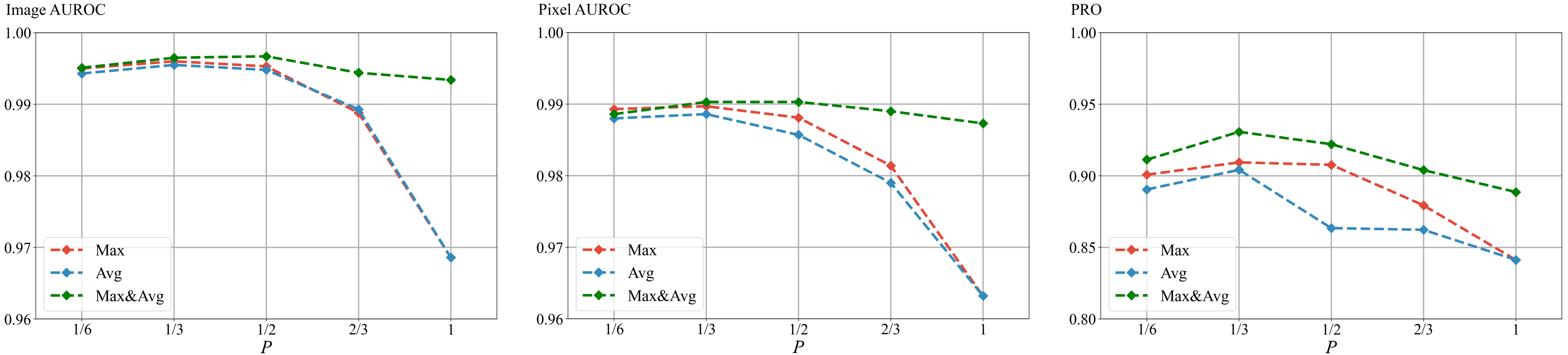}
  \caption{Performance of RealNet on MVTec-AD dataset \cite{bergmann2019mvtec} under various modes of reconstruction residuals selection (\texttt{Max}, \texttt{Avg}, and \texttt{Max}\&\texttt{Avg}) and varying retention ratio $P$ of reconstruction residuals.}
  \label{fig:fig5}
\end{figure*}

\subsection{Anomaly detection on other benchmarks}

To comprehensively assess the effectiveness of RealNet, we conduct experiments on the BTAD \cite{mishra2021vt} and VisA \cite{zou2022spot} datasets. On the VisA dataset \cite{zou2022spot}, characterized by complex structures and multiple detection objects, RealNet employing SIA yields a significant performance improvement, achieving an Image AUROC of 97.8\% and a Pixel AUROC of 98.8\%. In the case of the BTAD dataset \cite{mishra2021vt}, RealNet with SIA achieves competitive results, securing an Image AUROC of 96.1\% and a Pixel AUROC of 97.9\%. Detailed results can be found in \cref{sec: supC}.

\subsection{Ablation studies}

To evaluate the effectiveness of each module of RealNet, we conduct comprehensive ablation studies on MVTec-AD dataset \cite{bergmann2019mvtec}. First, we evaluate the impact of AFS and RRS on the performance of RealNet.

\textbf{W/O AFS:} We replace AFS with two alternative dimensionality reduction methods, namely Random Dimensionality Reduction (RDR) \cite{defard2021padim} and Random Linear Projections Reduction (RLPR) \cite{roth2022towards,xisoftpatch}. RDR randomly selects some dimensional features from high-dimensional features, while RLPR employs an untrained linear transformation layer for linear projection. We report the results of our RealNet with RDR and RLPR, respectively, as shown in the experiments 1 and 3 in \cref{tab:table5-a}. \textbf{W/O RRS:} We feed the global reconstruction residual $E(A_n)$ into the discriminator to generate anomaly scores, and the results are shown in the experiments 1 and 2 in \cref{tab:table5-a}.

As indicated by the ablation results in \cref{tab:table5-a}, RRS contributes significantly to performance improvement. Using all the reconstruction residuals to generate anomaly scores, reconstruction residuals lacking of anomaly information can lead to missed anomaly regions, resulting in a significant decrease in anomaly detection performance. Furthermore, AFS yields better anomaly detection results compared to RDR and RLPR. A straightforward visualization result about AFS is provided in \cref{sec: supD}.

We further investigate the impact of anomaly strength $s$ in SDAS, and the results are shown in \cref{tab:table5-b}. When $s$ equals 0, SDAS generates normal images in high probability density regions. Blending images may introduce false positive anomaly regions, which lowers the reconstruction difficulty and confuses the discriminator, leading to suboptimal performance. Conversely, when $s$ is too large, the synthetic anomaly images deviate from the true distribution of anomalous images, causing RealNet's performance to deteriorate. Our findings indicate that uniformly sampling $s$ within a specific range serves as a robust approach for generating anomalous images. This method enables RealNet to cover a wider range of anomalous patterns, ultimately improving the overall anomaly detection performance.

In \cref{fig:fig5}, we report the impact of different RRS modes and retention ratios on the performance of RealNet. Since the setting of $r$ is related to $m_k$, we introduce the retention ratio $P$, defined as: $P=\dfrac{r}{\Sigma_{k=1}^Km_k}$. Compared to \texttt{Max} and \texttt{Avg} modes, \texttt{Max}\&\texttt{Avg} mode demonstrates superior robustness in detecting anomalies across various scales. At equal retention rates, the \texttt{Max}\&\texttt{Avg} mode discards more reconstruction residuals lacking anomalous information than the \texttt{Max} and \texttt{Avg} modes, mitigating performance degradation and further emphasizing the effectiveness of the \texttt{Max}\&\texttt{Avg} mode in enhancing RealNet's anomaly detection capabilities. More ablation experiments and analysis can be found in \cref{sec: supC}.

\section{Conclusion}

In this work, we introduce RealNet, an innovative self-supervised anomaly detection framework. Our approach integrates three core components: Strength-controllable Diffusion Anomaly Synthesis (SDAS),  Anomaly-aware Features Selection (AFS), and Reconstruction Residuals Selection (RRS). These components synergistically contribute to RealNet, enabling effective exploitation of large-scale pre-trained models in anomaly detection while keeping the computational overhead within a reasonably low and acceptable range. RealNet provides a flexible foundation for future research in anomaly detection utilizing pre-trained feature reconstruction techniques. Through extensive experiments, we illustrate RealNet's proficiency in addressing diverse real-world anomaly detection challenges.

\noindent \textbf{Acknowledgements}. This work was supported in part by the National Natural Science Foundation of China under Grant 62177034 and Grant 61972046.

{
    \small
    \bibliographystyle{ieeenat_fullname}
    \bibliography{main}
}

\clearpage

\maketitlesupplementary
\renewcommand*{\thefigure}{S\arabic{figure}}
\renewcommand*{\thetable}{S\arabic{table}}
\renewcommand*{\theequation}{S\arabic{equation}}
\renewcommand*{\thealgorithm}{S\arabic{algorithm}}
\renewcommand\thesection{\Alph{section}}
\setcounter{table}{0}
\setcounter{figure}{0}
\setcounter{equation}{0}
\setcounter{section}{0}
\setcounter{algorithm}{0}

\appendix

\section{Overview}
\label{sec: supA}

We organize this supplementary material into the following sections: \cref{sec: supB} provides additional implementation details for RealNet. \cref{sec: supC} provides detailed results on the BTAD \cite{mishra2021vt} and VisA \cite{zou2022spot} datasets, supplementary ablation study results, an analysis of RealNet's computational efficiency, anomaly detection results in multi-class setting, as well as synthetic anomaly image quality assessment results. \cref{sec: supD} offers additional visualization results, including qualitative results of RealNet in anomaly localization, images generated by SDAS, and a straightforward visualization result of AFS. \cref{sec: supE} discusses the limitations of our method.

\section{More details}
\label{sec: supB}
In SDAS, we use the learnable reverse diffusion variance \cite{nichol2021improved} as $\Sigma_\theta(x_t,t)$, given by:

\begin{equation}
\label{eq:equs1}
\Sigma_\theta(x_t,t)=\exp(v\log\beta_t+(1-v)\log\tilde{\beta}_t)
\end{equation}
Here, $\beta_t$ represents the variance of the diffusion process, while $\tilde{\beta}_t$ represents the variance of the conditional posterior distribution $q(x_{t-1}|x_t,x_0)$, and $\tilde{\beta}_t=\frac{1-\bar{\alpha}_{t-1}}{1-\bar{\alpha}_{t}}\beta_t$. The vector $v$ is predicted by the model and weighted with $\beta_t$ and $\tilde{\beta}_t$ in the $log$ space. We optimize $\mu_\theta(x_t,t)$ and $\Sigma_\theta(x_t,t)$ with the loss $\mathcal{L}_{hybrid}$:
\begin{equation}
\label{eq:equs2}
\mathcal{L}_{hybrid}=\mathcal{L}_{simple}+\gamma\mathcal{L}_{vlb}
\end{equation}
where
\begin{gather}
\begin{align}
\mathcal{L}_{vlb}&=\mathcal{L}_0+\mathcal{L}_1+...+\mathcal{L}_{T-1}+\mathcal{L}_{T}\notag \\
\mathcal{L}_0&=-\log p_{\theta}(x_0|x_1)\notag \\
\mathcal{L}_{t-1}&=D_{KL}(q(x_{t-1}|x_t,x_0)||p_{\theta}(x_{t-1}|x_t))\notag \\
\mathcal{L}_{T}&=D_{KL}(q(x_{T}|x_0)||p(x_{T}))
\end{align}
\end{gather}

We set $\gamma$ to 0.001 in \cref{eq:equs2}, and stop the gradient of $\mu_\theta(x_t,t)$ in $\mathcal{L}_{vlb}$ during the training phase. To accelerate the convergence of the diffusion model, we initialize it with weights pre-trained on ImageNet \cite{deng2009imagenet}. We set the reverse diffusion step $T$ of 20, and generating 10,000 images at a resolution of $256 \times 256$ takes 6 hours using a single NVIDIA GeForce RTX 3090. 

The SDAS with DDIM \cite{song2020denoising} is described in \cref{alg:algs1}, which provides three options for applying perturbation variance in the deterministic reverse diffusion process: $\Sigma=\beta_t$, $\Sigma=\tilde{\beta}_{t}$, and $\Sigma=\Sigma_\theta(x_t,t)$. Experimental observations show that the anomaly images obtained by IDDPM \cite{nichol2021improved} are slightly better than those obtained by DDIM \cite{song2020denoising}, and therefore, we use IDDPM \cite{nichol2021improved} for SDAS. Some examples can be found in \cref{sec: supD}.

\begin{figure}[t]
  \centering
  \includegraphics[width=\linewidth]{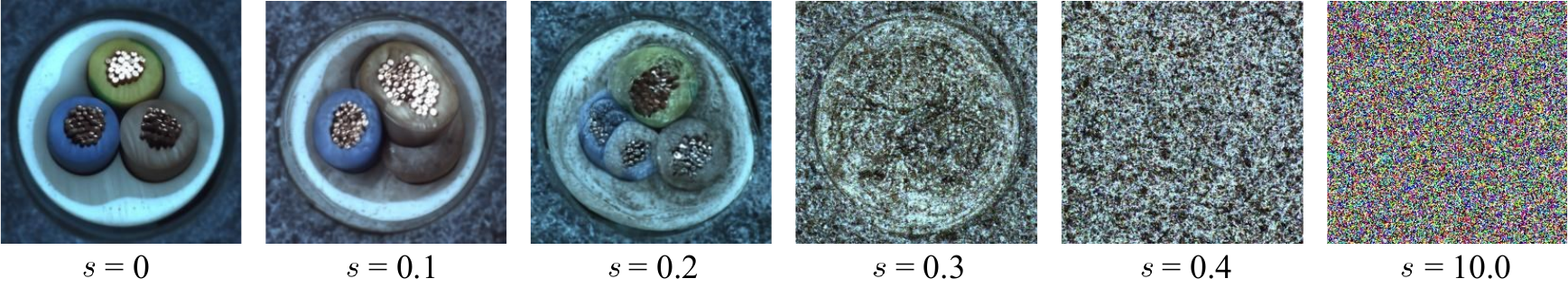}
  \caption{Sample anomaly images generated by SDAS with different anomaly strengths $s$.}
  \label{fig:figs1}
  \vspace{-0.4cm}
\end{figure}

\cref{fig:figs1} presents examples of images generated by SDAS with a broader range of anomaly strengths. As the anomaly strength increases, the generated anomalous images contain more noise, reducing their authenticity. In the experiments, we set the anomaly strength between 0.1 and 0.2, allowing SDAS to encompass a wider range of real-world anomalies.

In RRS, the global reconstruction residual $E(A_{n})$ originates from distinct reconstruction networks, leading to disparate distributions across its dimensions. We apply a \texttt{BatchNorm} \cite{ioffe2015batch} layer (without Affine) to $E(A_{n})$ and then perform reconstruction residuals selection to ensure a consistent distribution across the dimensions of $E(A_{n})$. 

The discriminator is implemented using a basic MLP with upsampling layers to map anomaly scores from feature resolution to image resolution. During the training phase of RealNet, we do not use any data augmentation for the synthesis of anomalous images, and maintain an equal ratio between normal images and synthetic anomalous images. In the process of image blending, we uniformly sample the opacity $\delta$ from 0.5 to 1.0 in Eq. (\textcolor{red}{3}). The training of RealNet is performed on a single NVIDIA GeForce RTX 3090, with an approximate average training time of 2 hours. 

\begin{algorithm*}[t]
\caption{SDAS with DDIM \cite{song2020denoising}}
\begin{algorithmic}
\STATE \textbf{Input}: \text{diffusion model $\epsilon_\theta(x_t,t)$, perturbation variance $\Sigma$, anomaly strength $s$}
\STATE $x_T \sim \mathcal N(0,\textbf{I})$
\STATE \textbf{for all} $t$ \text{from T to 1} \textbf{do}
\STATE \hspace{0.4cm}$x_{t-1} \sim \mathcal N(\sqrt{\bar{\alpha}_{t-1}}(\frac{x_t-\sqrt{1-\bar{\alpha}_{t}}\epsilon_\theta(x_t,t) }{\sqrt{\bar{\alpha}_{t}}})+\sqrt{1-\bar{\alpha}_{t-1}}\epsilon_\theta(x_t,t),s\Sigma)$
\STATE \textbf{end for}
\STATE \textbf{return} $x_0$
\end{algorithmic}
\label{alg:algs1}
\end{algorithm*}

\begin{figure*}[t]
\begin{minipage}{\textwidth}
    \centering
    \makeatletter\def\@captype{table}\makeatother
    \renewcommand\arraystretch{1.1}
    \caption{Comparison of RealNet with alternative anomaly detection methods on the BTAD dataset \cite{mishra2021vt}, employing Image AUROC (\%) and Pixel AUROC (\%) as evaluation metrics.}
    \label{tab:tables1}
    \resizebox{\linewidth}{!}{
    \tiny
    \begin{tabular}{c|ccccc|cc}
    \bottomrule
    \multicolumn{1}{c|}{Category} & VT-ADL \cite{mishra2021vt} & P-SVDD \cite{yi2020patch} & FastFlow \cite{yu2021fastflow} & SPADE \cite{cohen2020sub} & RD++ \cite{tien2023revisiting} & RealNet (SIA) &  RealNet (DTD \cite{cimpoi2014describing})\\
    \hline
    \multicolumn{1}{c|}{01} & (-, \textbf{99}) & (95.7, 91.6) & (-, 95) & (91.4, 97.3) &  (96.8, 96.2) &    (\textbf{100.0}, 98.2)        & (\textbf{100.0}, 98.1)\\
    \multicolumn{1}{c|}{02} & (-, 94) & (72.1, 93.6) & (-, 96) & (71.4, 94.4) &      (\textbf{90.1}, \textbf{96.4})      &   (88.6, 96.3)& (87.5, 96.3)\\
    \multicolumn{1}{c|}{03} & (-, 77) & (82.1, 91.0) & (-, 99) & (99.9, 99.1) &   (\textbf{100.0}, \textbf{99.7}) &  (99.6, 99.4)           & (99.4, 99.6)\\
    \textbf{AVG} & (-, 90.0) & (83.3, 92.1) & (-, 96.7) & (87.6, 96.9) &         (95.6,97.4)           &  (\textbf{96.1}, 97.9)         &  (95.7, \textbf{98.0})\\
    \toprule
    \end{tabular}%
  }

\end{minipage}\quad
\begin{minipage}{\textwidth}
    \centering
    \makeatletter\def\@captype{table}\makeatother
    \renewcommand\arraystretch{1.1}
    \caption{Comparison of RealNet with alternative anomaly detection methods on the VisA dataset \cite{zou2022spot}, employing Image AUROC (\%) and Pixel AUROC (\%) as evaluation metrics.}
    \label{tab:tables2}
    \resizebox{\linewidth}{!}{
      \tiny
    \begin{tabular}{c|cccc|cc}
    \bottomrule
    Category & SPADE \cite{cohen2020sub} & FastFlow \cite{yu2021fastflow} & DRAEM \cite{zavrtanik2021draem} & PatchCore \cite{roth2022towards} & RealNet (SIA) & RealNet (DTD \cite{cimpoi2014describing}) \\
    \hline
    Candle & (91.0, 97.9) & (92.8, 94.9) & (91.8, 96.6) & (\textbf{98.6}, \textbf{99.5}) & (96.1, 99.1) & (95.0, 99.0) \\
    Capsules & (61.4, 60.7) & (71.2, 75.3) & (74.7, 98.5) & (81.6, \textbf{99.5}) & (\textbf{93.2}, 98.7) & (88.1, 97.6) \\
    Cashew & (\textbf{97.8}, 86.4) & (91.0, 91.4) & (95.1, 83.5) & (97.3, \textbf{98.9}) & (\textbf{97.8}, 98.3) & (95.9, 97.6) \\
    Chewing gum & (85.8, 98.6) & (91.4, 98.6) & (94.8, 96.8) & (99.1, 99.1) & (99.9, \textbf{99.8}) & (\textbf{100.0}, \textbf{99.8}) \\
    Fryum & (88.6, 96.7) & (88.6, \textbf{97.3}) & (\textbf{97.4}, 87.2) & (96.2, 93.8) & (97.1, 96.2) & (95.3, 95.2) \\
    Macaroni1 & (95.2, 96.2) & (98.3, 97.3) & (97.2, \textbf{99.9}) & (97.5, 99.8) & (\textbf{99.8}, \textbf{99.9}) & (98.2, 99.7) \\
    Macaroni2 & (87.9, 87.5) & (86.3, 89.2) & (85.0, 99.2) & (78.1, 99.1) & (\textbf{95.2}, \textbf{99.6}) & (91.8, 99.3) \\
    PCB1  & (72.1, 66.9) & (77.4, 75.2) & (47.6, 88.7) & (\textbf{98.5}, \textbf{99.9}) & (\textbf{98.5}, 99.7) & (97.1, 99.4) \\
    PCB2  & (50.7, 71.1) & (61.9, 67.3) & (89.8, 91.3) & (97.3, \textbf{99.0}) & (\textbf{97.6}, 98.0) & (97.5, 97.8) \\
    PCB3  & (90.5, 95.1) & (74.3, 94.8) & (92.0, 98.0) & (97.9, \textbf{99.2}) & (\textbf{99.1}, 98.8) & (97.6, 98.4) \\
    PCB4  & (83.1, 89.0) & (80.9, 89.9) & (98.6, 96.8) & (99.6, \textbf{98.6}) & (\textbf{99.7}, \textbf{98.6}) & (99.2, \textbf{98.6}) \\
    Pipe fryum & (81.1, 81.8) & (72.0, 87.3) & (\textbf{100.0}, 85.8) & (99.8, 99.1) & (99.9, \textbf{99.2}) & (99.9, 98.6) \\
    \hline
    \textbf{AVG}   & (82.1, 85.6) & (82.2, 88.2) & (88.7, 93.5) & (95.1, \textbf{98.8}) & (\textbf{97.8}, \textbf{98.8}) & (96.3, 98.4) \\
    \toprule
    \end{tabular}%
    }

\end{minipage}\quad
\end{figure*}

\section{More results}
\label{sec: supC}

\subsection{Experimental results on BTAD}

We evaluate the anomaly detection and localization performance of RealNet and alternative methods on the BTAD dataset \cite{mishra2021vt}, with the results shown in \cref{tab:tables1}. Although SIA does not show a significant performance improvement compared to DTD \cite{cimpoi2014describing} due to the absence of complex structural anomalies in the three industrial products of the BTAD dataset \cite{mishra2021vt}, RealNet demonstrates state-of-the-art performance in anomaly detection and localization when compared to other methods, without any structural or hyperparameter tuning.

\subsection{Experimental results on VisA}

We present the performance of RealNet and alternative methods on the VisA dataset under the one-class protocol \cite{zou2022spot} in \cref{tab:tables2}. RealNet achieves the best performance in both anomaly detection and localization. Compared to DTD \cite{cimpoi2014describing}, the RealNet trained using SIA shows an improvement of 1.5\% in Image AUROC and 0.4\% in Pixel AUROC.

\subsection{Supplementary ablation studies}

\begin{figure*}[t]
\begin{minipage}{0.98\textwidth}
    \centering
    \makeatletter\def\@captype{table}\makeatother
    \renewcommand\arraystretch{1.13}
    \caption{Performance evaluation of RealNet with varying backbones and reconstruction feature dimension settings on the MVTec-AD dataset \cite{bergmann2019mvtec}, employing Image AUROC (\%), Pixel AUROC (\%), and PRO (\%) as evaluation metrics.}
    \label{tab:tables3}
    \resizebox{\linewidth}{!}{
    \scriptsize
     \begin{tabular}{c|c|c|cc}
    \bottomrule
    Backbone & EfficientNetB4 \cite{tan2019efficientnet} & ResNet34 \cite{he2016deep} & \multicolumn{2}{c}{WideResNet50 \cite{zagoruyko2016wide}} \\
    \hline
    \multicolumn{1}{c|}{\{$m_1$,...,$m_{K}$\}} & \{24, 32, 56, 160\} & \{64, 128, 256, 128\} & \{128, 256, 256, 128\} & \{256, 512, 512, 256\} \\
    \hline
    Bottle & (\textbf{100.0}, 98.83, \textbf{95.96}) & (\textbf{100.0}, 98.56, 95.91) & (\textbf{100.0}, \textbf{99.41}, 94.37) & (\textbf{100.0}, 99.30, 95.62) \\
    Cable & (96.36, 96.33, 88.61) & (96.31, 96.32, 88.68) & (98.35, 98.01, 92.99) & (\textbf{99.19}, \textbf{98.10}, \textbf{93.38}) \\
    Capsule & (97.97, 99.16, \textbf{91.46}) & (96.81, 98.78, 87.87) & (99.44, \textbf{99.39}, 79.76) & (\textbf{99.56}, 99.32, 84.48) \\
    Carpet & (\textbf{100.0}, 98.27, 96.35) & (99.76, 98.37, 94.45) & (99.80, 98.91, 96.32) & (99.84, \textbf{99.19}, \textbf{96.41}) \\
    Grid  & (99.92, 99.31, 97.35) & (\textbf{100.0}, 99.26, \textbf{97.39}) & (\textbf{100.0}, \textbf{99.55}, 96.38) & (\textbf{100.0}, 99.51, 97.28) \\
    Hazelnut & (99.89, 98.45, \textbf{94.98}) & (99.93, 99.35, 94.36) & (\textbf{100.0}, 99.67, 93.06) & (\textbf{100.0}, \textbf{99.68}, 93.14) \\
    Leather & (\textbf{100.0}, 99.34, 97.75) & (99.97, 99.40, \textbf{98.28}) & (\textbf{100.0}, \textbf{99.81}, 96.99) & (\textbf{100.0}, 99.76, 96.22) \\
    Metal Nut & (99.07, 96.90, 92.65) & (99.17, 96.68, 93.34) & (\textbf{99.90}, \textbf{98.75}, \textbf{95.10}) & (99.76, 98.58, 94.39) \\
    Pill  & (96.10, 94.86, 86.60) & (97.55, 98.23, \textbf{93.17}) & (97.85, \textbf{99.19}, 80.73) & (\textbf{99.13}, 99.02, 91.04) \\
    Screw & (92.95, 99.05, \textbf{92.68}) & (96.99, 99.09, 89.57) & (97.99, 99.28, 88.60) & (\textbf{98.83}, \textbf{99.45}, 87.90) \\
    Tile  & (99.49, 95.69, 92.10) & (99.93, 97.40, 91.65) & (\textbf{100.0}, 99.27, 97.20) & (99.96, \textbf{99.44}, \textbf{97.70}) \\
    Toothbrush & (99.44, 98.90, \textbf{92.39}) & (\textbf{100.0}, 98.26, 91.74) & (\textbf{100.0}, \textbf{99.26}, 91.22) & (99.44, 98.71, 91.57) \\
    Transistor & (99.58, \textbf{98.57}, \textbf{93.63}) & (99.33, 97.70, 88.53) & (99.79, 98.26, 83.34) & (\textbf{100.0}, 98.00, 92.92) \\
    Wood  & (98.77, 94.47,\textbf{92.67}) & (98.16, 96.35, 91.46) & (\textbf{99.56}, \textbf{98.22}, 90.76) & (99.21, \textbf{98.22}, 90.54) \\
    Zipper & (99.71, 98.01, 91.68) & (\textbf{99.90}, 98.55, \textbf{93.91}) & (99.74, \textbf{99.20}, 90.73) & (99.82, 99.17, 93.43) \\
    \hline
    \textbf{AVG}   & (98.62, 97.74, \textbf{93.12}) & (98.92, 98.15, 92.69) & (99.49, \textbf{99.07}, 91.17) & (\textbf{99.65}, 99.03, 93.07) \\
    \toprule
    \end{tabular}
  }
\end{minipage}\quad
\begin{minipage}{\textwidth}
    \centering
    \makeatletter\def\@captype{figure}\makeatother
    \includegraphics[width=0.98\textwidth]{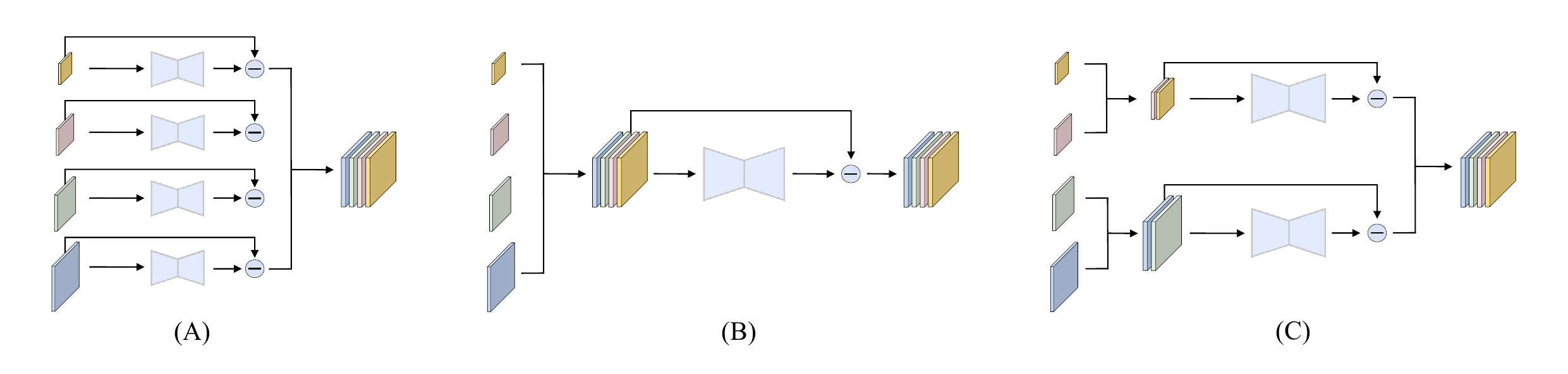}
    \caption{Various architectures of multi-scale feature reconstruction for anomaly detection. (A) Independent Reconstruction Architecture uses separate networks for multi-scale feature reconstruction. (B) Fully Aligned Feature Reconstruction Architecture aligns all features for reconstruction. (C) Neighboring Aligned Feature Reconstruction Architecture aligns and reconstructs neighboring resolution features. }
    \label{fig:figs2}
\end{minipage}
\end{figure*}

To further investigate RealNet's anomaly detection performance on the MVTec-AD dataset \cite{bergmann2019mvtec}, we examine various backbones and reconstruction feature dimension settings. As shown in \cref{tab:tables3}, when WideResNet50 \cite{zagoruyko2016wide} is employed as the backbone and the reconstruction feature dimensions $\{m_1,...,m_K\}$ are reduced from \{256, 512, 512, 256\} to \{128, 256, 256, 128\}, there is a slight decrease of 0.16\% in Image AUROC. Despite this reduction, RealNet maintains its competitive performance compared to other methods. Additionally, the adoption of EfficientNetB4 \cite{tan2019efficientnet} and ResNet34 \cite{he2016deep} as backbones also results in competitive performance, demonstrating the effectiveness of RealNet across various settings.

\subsection{Computational efficiency analysis}

We investigate the computational efficiency and detection performance of three different multi-scale feature reconstruction architectures on the MVTec-AD dataset \cite{bergmann2019mvtec}, as illustrated in  \cref{fig:figs2}. To provide a comprehensive analysis, \cref{tab:tables4} presents the inference speed, model size (including backbone), and anomaly detection performance of these architectures. The inference is performed using a single Nvidia GeForce RTX 3090, with all other settings adhering to the specifications detailed in Sec. \textcolor{red}{4.1}.

We utilize a consistent reconstruction network based on the U-Net model with skip connections across three distinct architectures. The employed U-Net model initiates with a stack of residual layers and down-sampling layers, gradually decreasing the spatial dimensions while increasing the number of channels. Subsequently, the model utilizes a stack of residual layers and up-sampling layers to inversely reconstruct features. Throughout this process, skip connections are incorporated at equivalent spatial resolutions to ensure a smooth and logical flow.

Specifically, architecture \textbf{A} adopts separate reconstruction networks to reconstruct multi-scale features without the need for feature interpolation or alignment. This method ensures outstanding anomaly detection performance while maintaining high computational efficiency. With a resolution of $256 \times 256$ and reconstruction feature dimensions of \{256, 512, 512, 256\}, architecture \textbf{A} with model size of 2.2 GB achieves a rapid inference speed of 31.93 FPS. And it can perform inference using only 4GB of GPU memory. Concurrently, it attains an Image AUROC of 99.65\% and a Pixel AUROC of 99.03\%. By decreasing the reconstruction feature dimensions to \{128, 256, 256, 128\}, architecture \textbf{A} reduces the model size to 0.74 GB and achieves a higher inference speed of 40.42 FPS, while preserving an Image AUROC of 99.49\% and a Pixel AUROC of 99.07\%. Furthermore, at a high resolution of $512 \times 512$, it delivers an inference speed of 13.53 FPS, along with an Image AUROC of 99.40\% and a Pixel AUROC of 98.71\%. These inference speeds indicate that architecture \textbf{A} satisfies the real-time requirements for industrial inspection applications.

Regarding architecture \textbf{B}, as referenced in \cite{yang2020dfr,tao2022unsupervised,you2022unified}, it is used to align the multi-scale features of a small pre-trained network. As aligning down-sampled features will reduce the resolution of model detection and cause predictable performance loss, the experiment only discusses up-sampling alignment. Compared to architecture \textbf{A}, architecture \textbf{B} reconstructs the interpolated features, significantly reducing computational efficiency and increasing model size. Moreover, due to the limited number of normal images, the overly large reconstruction network in architecture \textbf{B} is prone to overfitting, resulting in reduced detection performance. Consequently, for large-scale pre-trained networks with high-dimensional features, aligning and reconstructing all features is suboptimal.

Moreover, we observe that utilizing multiple reconstruction networks for feature reconstruction in architecture \textbf{A} causes minor deviations in localizing small-area anomalies, resulting in a reduced PRO. To address this, we propose architecture \textbf{C}, which aligns and reconstructs features from two neighboring resolution, thereby reducing the number of reconstruction networks, controlling the model size, and striking a balance between computational efficiency and localization accuracy. At a $256 \times 256$ resolution, with reconstruction feature dimensions of \{256, 512, 512, 256\}, architecture \textbf{C} has a 3.75 GB model size and achieves an inference speed of 22.39 FPS, while attaining an Image AUROC of 99.62\%, a Pixel AUROC of 98.90\%, and a PRO of 94.71\%.

In summary, the design of RealNet balances both anomaly detection performance and computational efficiency. The introduction of AFS allows us to flexibly customize models of various sizes to accommodate different usage scenarios. Furthermore, among our three key innovations, both AFS and RRS introduce no additional learnable parameters, ensuring strong interpretability. As for SDAS, it only introduces perturbation during the reverse diffusion process, without requiring any prior knowledge about the distribution of real anomaly images.

\begin{table}[t]
  \centering
  \renewcommand\arraystretch{1.2}
  \caption{Performance evaluation of various reconstruction architectures on the MVTec-AD dataset \cite{bergmann2019mvtec}. The metrics include Image AUROC (\%), Pixel AUROC (\%), and PRO (\%).}
  \resizebox{\linewidth}{!}{
    	  \begin{tabular}{c|c|c|c}
            \bottomrule
            \multicolumn{1}{c|}{} & \multicolumn{1}{c|}{Speed (FPS) $\uparrow$} & \multicolumn{1}{c|}{Model Size (GB) $\downarrow$} & Metrics $\uparrow$\\
            \toprule
            \multicolumn{3}{c}{\{$m_1$,...,$m_{K}$\} is \{128, 256, 256, 128\} and image size is $256 \times 256$}&\multicolumn{1}{c}{}  \\
             \bottomrule
            A     & \textbf{40.42} & \textbf{0.74} & (99.49, \textbf{99.07}, 91.17) \\
            \toprule
            \multicolumn{3}{c}{\{$m_1$,...,$m_{K}$\} is \{256, 512, 512, 256\} and image size is $256 \times 256$}&\multicolumn{1}{c}{}  \\
            \bottomrule
            A     & 31.93 & 2.20 & (\textbf{99.65}, 99.03, 93.07) \\
            B     & 10.83 & 7.22  & (98.44, 98.17, 94.27) \\
            C     & 22.39 & 3.75  & (99.62, 98.90, \textbf{94.71}) \\
            \toprule
            \multicolumn{3}{c}{\{$m_1$,...,$m_{K}$\} is \{256, 512, 512, 256\} and image size is $512 \times 512$}&\multicolumn{1}{c}{}  \\
            \bottomrule
            A     & 13.53 & 2.20 & (99.40, 98.71, 94.01) \\
            \toprule
            \end{tabular}%
        }
  \label{tab:tables4}%
\end{table}%

\subsection{Anomaly detection in multi-class setting}

In the multi-class setting \cite{you2022unified,zhao2023omnial}, anomaly detection is performed across multiple target classes concurrently, without access to sample class labels during both training and inference phases. Learning the data distributions of multiple classes jointly makes the reconstruction more complex. In such settings, previous reconstruction methods tend to output copies of the input images instead of performing selective reconstruction, which leads to a significant decrease in performance. We evaluate the performance of RealNet in multi-class anomaly detection on the MVTec-AD dataset \cite{bergmann2019mvtec} and compare it with alternative state-of-the-art methods. We use DTD \cite{cimpoi2014describing} for anomaly synthesis as class labels are unavailable during training. The remaining settings are consistent with Sec. \textcolor{red}{4.1}.

The results are shown in Tab. \ref{tab:tables5}. When detecting anomalies across 15 categories of the MVTec-AD dataset \cite{bergmann2019mvtec} concurrently, RealNet achieves an Image AUROC of 97.3\% and a Pixel AUROC of 98.4\% using a ResNet50 \cite{he2016deep} pre-trained on ImageNet \cite{deng2009imagenet}, surpassing state-of-the-art multi-class anomaly detection methods \cite{you2022unified,zhao2023omnial}. To ensure that normal regions can be reconstructed correctly, we do not explicitly constrain the generalization ability of the reconstructed network in RealNet. Instead, we implicitly constrain the reconstruction network to ensure that anomalous regions can be correctly detected by discarding a part of the reconstruction residuals.

\begin{table}[t]
  \centering
  \renewcommand\arraystretch{1.12}
  \caption{Comparison of RealNet with alternative methods in multi-class anomaly detection on the MVTec-AD dataset \cite{bergmann2019mvtec}.}
   \resizebox{0.91\linewidth}{!}{
        \tiny
        \begin{tabular}{c|cc}
        \bottomrule
        Methods & \multicolumn{1}{c}{Image AUROC} & \multicolumn{1}{c}{Pixel AUROC} \\
        \hline
        DRAEM \cite{zavrtanik2021draem} & 88.1  & 87.2 \\
        PaDiM \cite{defard2021padim} & 84.2  & 89.5 \\
        UniAD \cite{you2022unified} & 96.5  & 96.8 \\
        OmniAL \cite{zhao2023omnial} & 97.2  & 98.3 \\
        \hline
        RealNet & \textbf{97.3} & \textbf{98.4} \\
        \toprule
        \end{tabular}%
    }
  \label{tab:tables5}%
\end{table}%

\begin{table}[t]
  \centering
  \renewcommand\arraystretch{1.12}
  \caption{Image quality comparison of SIA with alternative anomaly synthesis approaches on the MVTec-AD dataset \cite{bergmann2019mvtec}.}
   \resizebox{0.91\linewidth}{!}{
        \tiny
        \begin{tabular}{c|cc}
        \bottomrule
        Methods & \multicolumn{1}{c}{FID \cite{heusel2017gans} $\downarrow$ } & \multicolumn{1}{c}{LPIPS \cite{zhang2018unreasonable} $\uparrow$ } \\
        \hline
        DTD \cite{cimpoi2014describing} & 120.52$\pm$0.63  & 0.16$\pm$0.00 \\
        CutPaste \cite{li2021cutpaste} & 77.34$\pm$0.09  & 0.11$\pm$0.00 \\
        NSA \cite{schluter2022natural} & 68.76$\pm$0.16  & 0.09$\pm$0.01 \\
        \hline
        SIA & \textbf{60.39$\pm$1.26} & \textbf{0.18$\pm$0.01} \\
        \toprule
        \end{tabular}%
    }
  \label{tab:tables6}%
\end{table}%

\subsection{Synthetic anomaly image quality assessment}

In this section, we evaluate the quality of anomaly images generated by various anomaly synthesis methods on the MVTec-AD dataset \cite{bergmann2019mvtec}. Specifically, we use the following evaluation metrics: 
\begin{itemize}
\item FID (Fr{\'e}chet Inception Distance) \cite{heusel2017gans}: FID measures the distance between the distribution of synthetic anomaly images and real anomaly images, evaluating both the realism and diversity of the synthetic anomaly images. A lower value indicates better performance.
\item LPIPS (Learned Perceptual Image Patch Similarity) \cite{zhang2018unreasonable}: We employ cluster-based LPIPS \cite{duan2023few} to evaluate the diversity of synthetic anomaly images. Supposing a category contains $N$ real anomaly images, we partition the synthesized anomaly images into $N$ groups by finding the lowest LPIPS, then we compute the mean pairwise LPIPS within each group and compute the average of all groups. A higher cluster LPIPS indicates greater diversity.
\end{itemize}

We employ various anomaly synthesis methods to generate 1,000 anomaly images for evaluation, with each method independently assessed three times. The experimental results are shown in \cref{tab:tables6}. In comparison to other anomaly synthesis methods, SIA achieves the best FID and LPIPS metrics, highlighting the outstanding performance of SDAS in generating both realistic and diverse anomaly images, and demonstrating the effectiveness of SDAS in improving anomaly detection performance.

\section{Visualization}

\label{sec: supD}
\begin{figure*}[htbp]
  \centering
  \includegraphics[width=\textwidth]{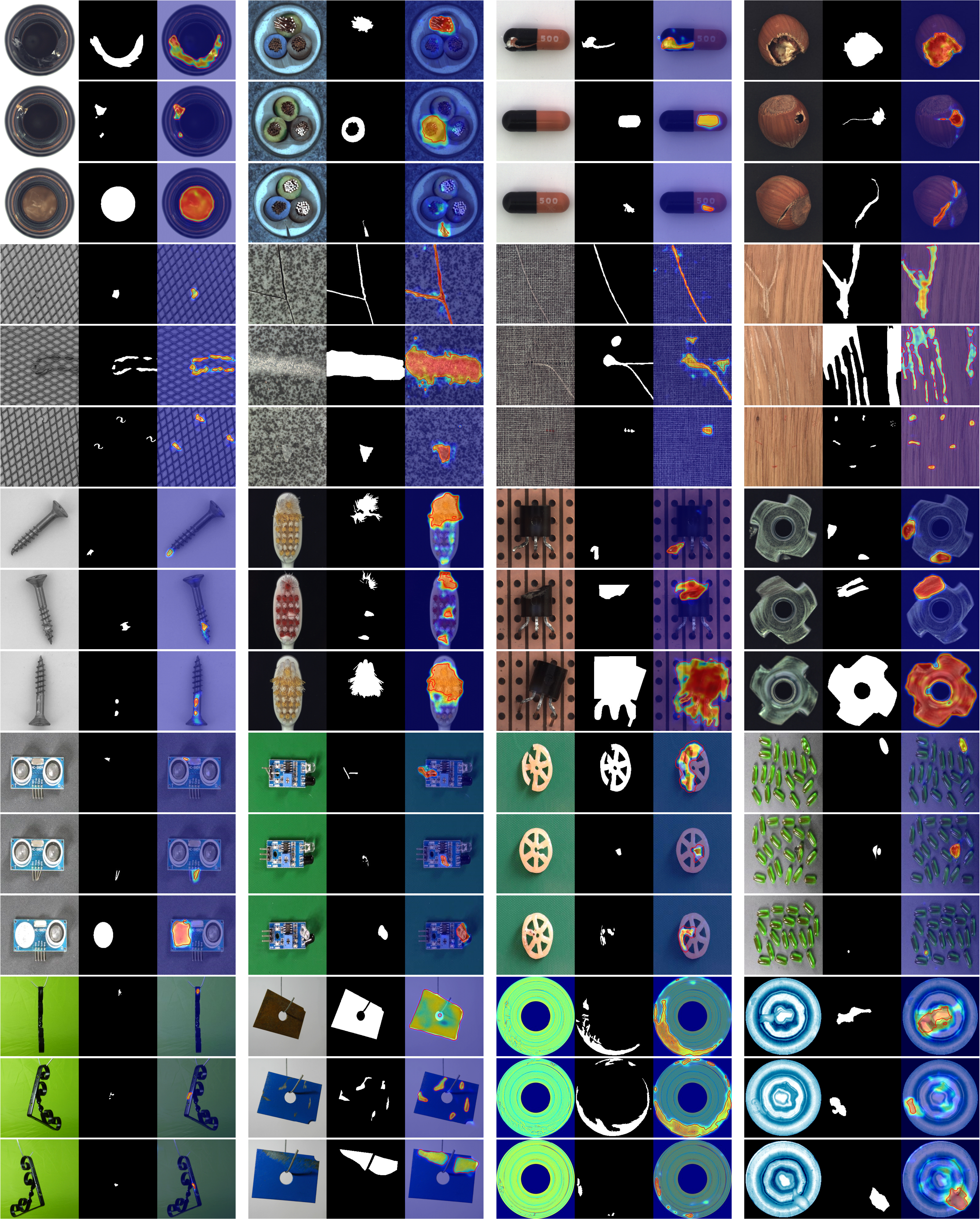}
  \caption{Qualitative results of RealNet. Within each group, from left to right, are the anomaly image, ground-truth, and predicted anomaly score. The examples are from the MVTec-AD \cite{bergmann2019mvtec}, MPDD \cite{jezek2021deep}, BTAD \cite{mishra2021vt}, and VisA \cite{zou2022spot} datasets.}
  \label{fig:figs3}
\end{figure*}

\begin{figure*}[htbp]
  \centering
  \includegraphics[width=\textwidth]{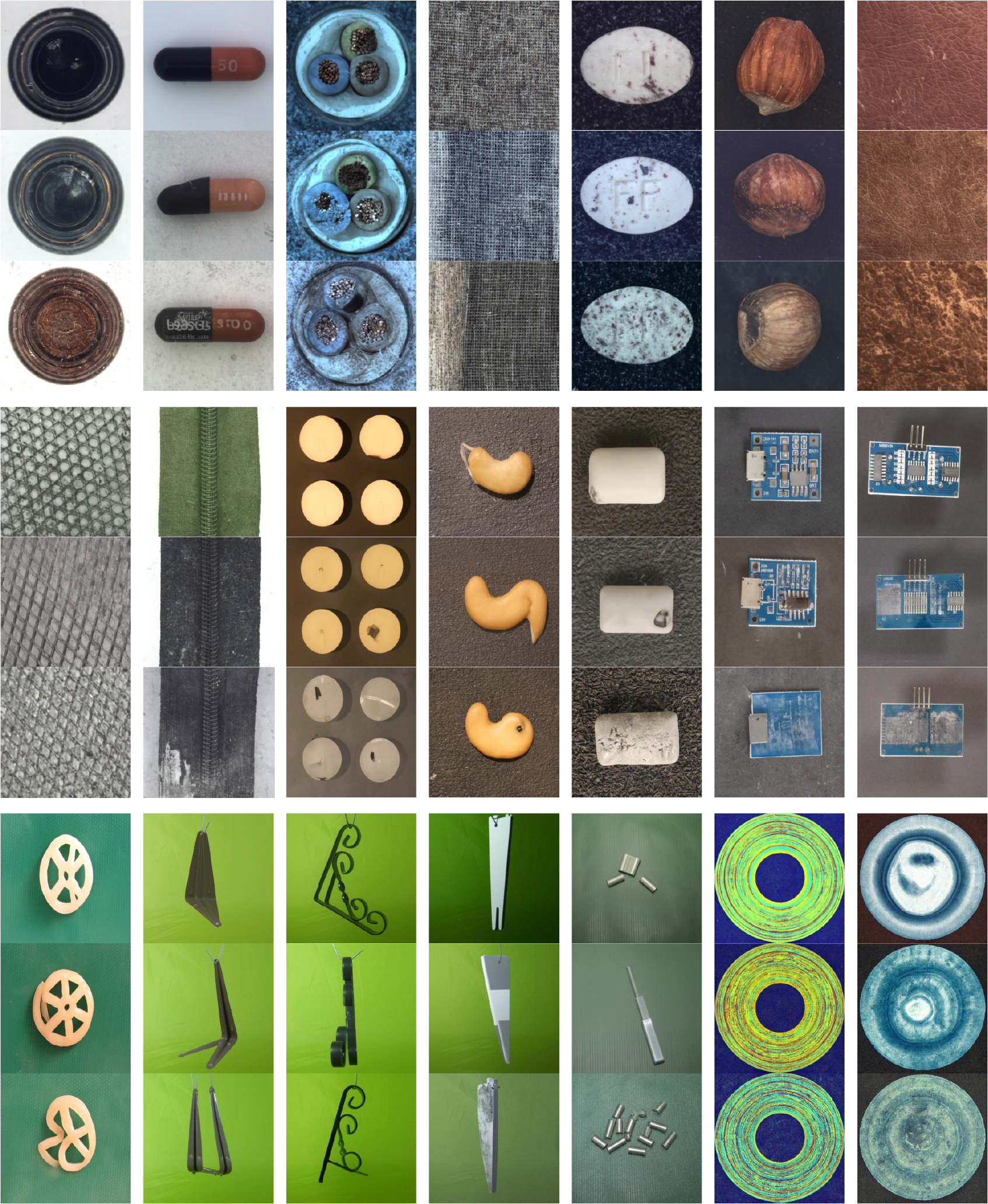}
  \caption{Anomaly images generated by SDAS. The examples are from the MVTec-AD \cite{bergmann2019mvtec}, MPDD \cite{jezek2021deep}, BTAD \cite{mishra2021vt}, and VisA \cite{zou2022spot} datasets. Within each group, from top to bottom, the anomaly strength gradually increases.}
  \label{fig:figs4}
\end{figure*}

\begin{figure*}[htbp]
  \centering
  \includegraphics[width=\textwidth]{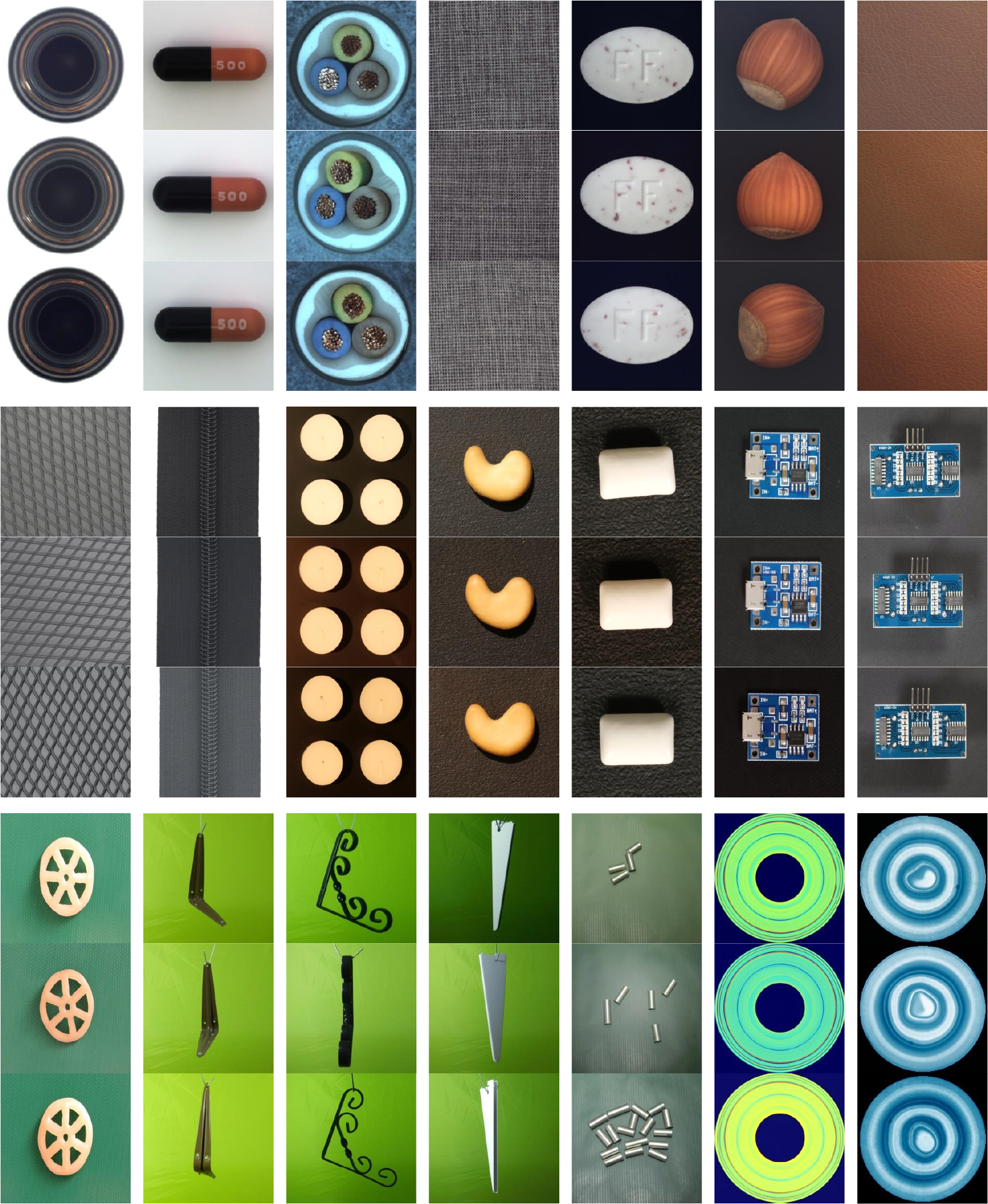}
  \caption{Normal images generated by SDAS (when $s=0$). The examples are from the MVTec-AD \cite{bergmann2019mvtec}, MPDD \cite{jezek2021deep}, BTAD \cite{mishra2021vt}, and VisA \cite{zou2022spot} datasets.}
  \label{fig:figs5}
\end{figure*}

\begin{figure*}[htbp]
  \centering
  \includegraphics[width=\textwidth]{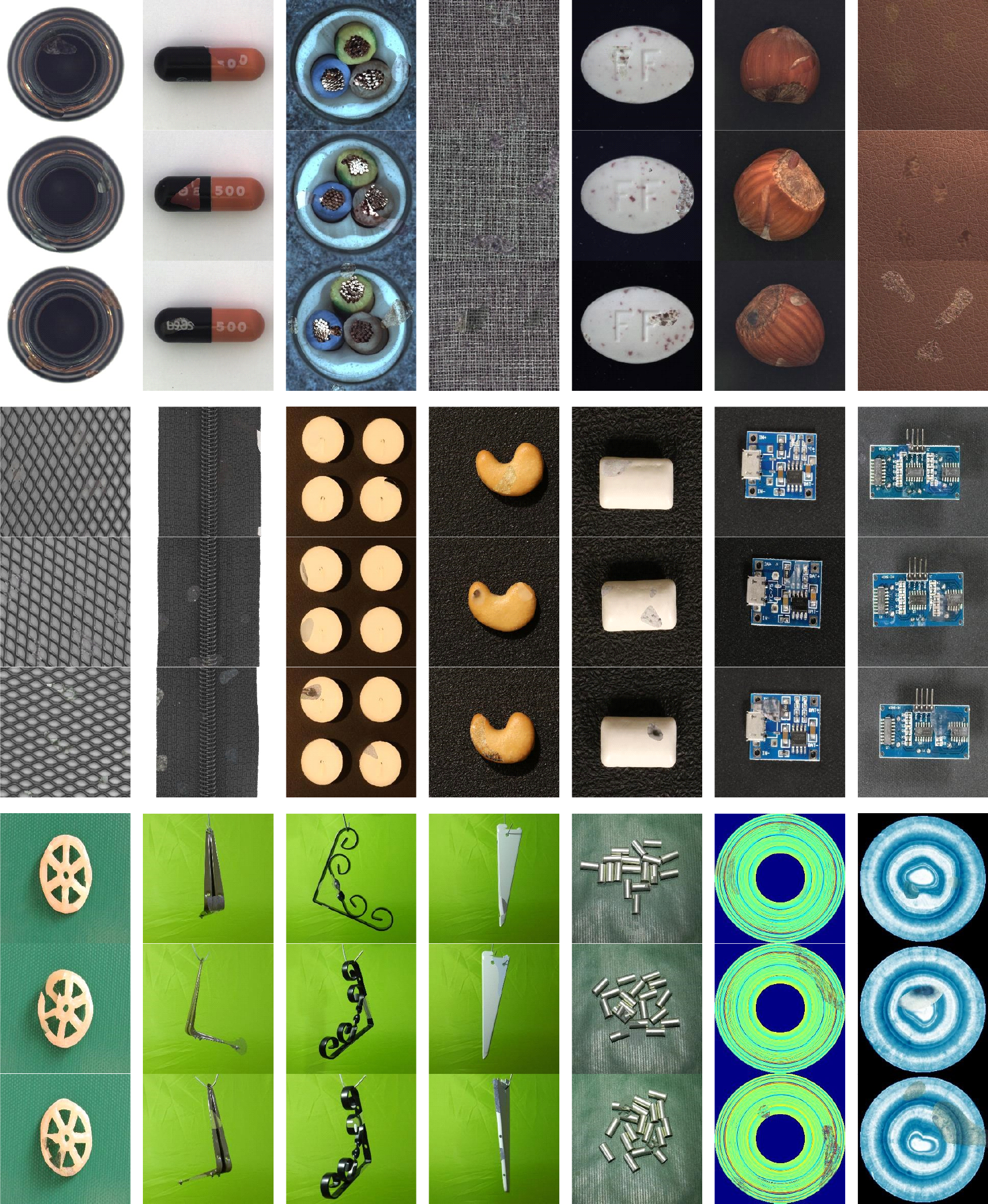}
  \caption{Local anomaly images synthesized by SIA. The examples are from the MVTec-AD \cite{bergmann2019mvtec}, MPDD \cite{jezek2021deep}, BTAD \cite{mishra2021vt}, and VisA \cite{zou2022spot} datasets. Within each group, from top to bottom, the anomaly strength gradually increases.}
  \label{fig:figs6}
\end{figure*}

\begin{figure*}[htbp]
  \centering
  \includegraphics[width=0.95\textwidth]{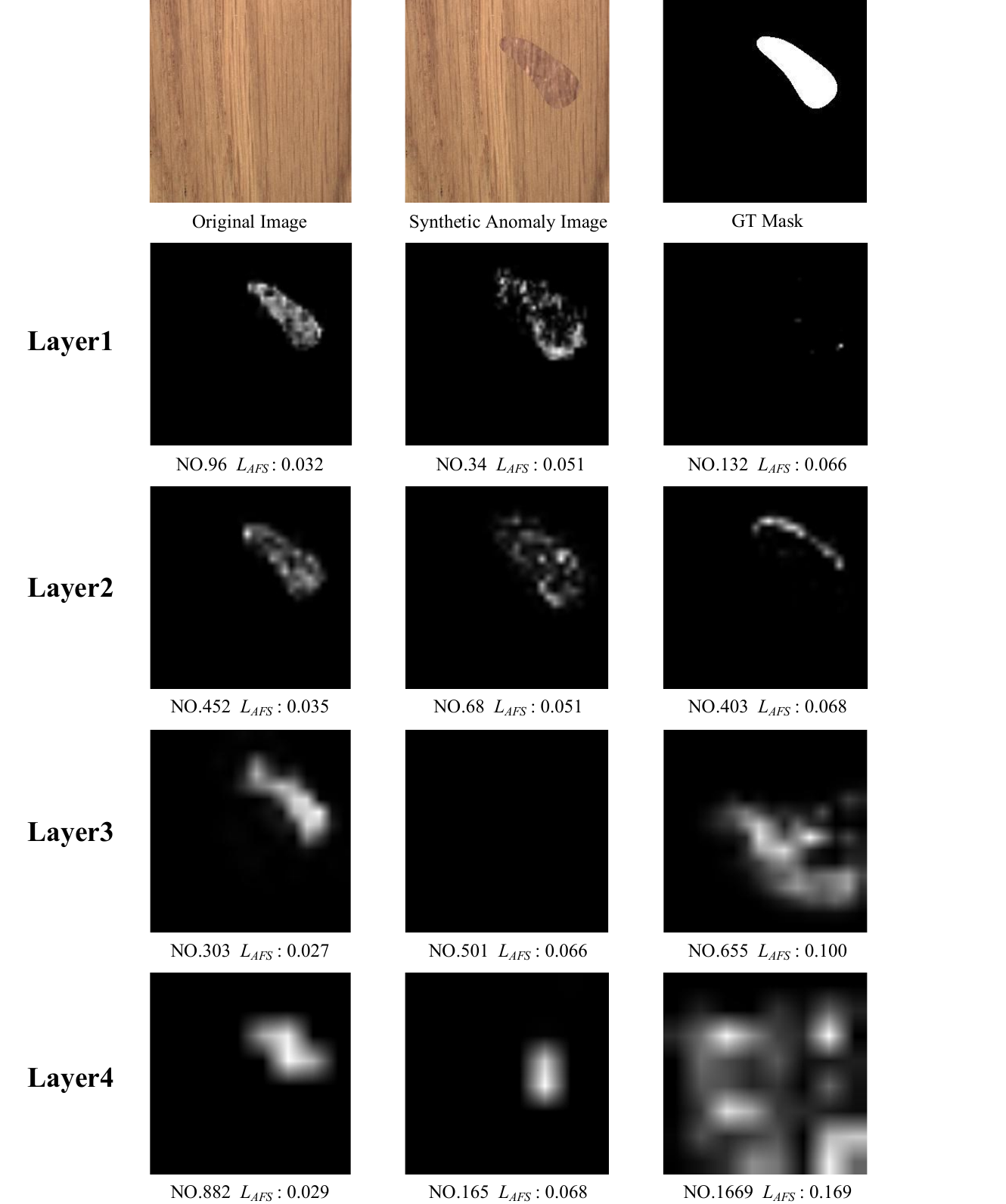}
  \caption{Visualization of AFS. For an original image and a synthetic anomaly image, we visualize the normalized difference between their corresponding feature maps across different layers of a pre-trained WideResNet50 \cite{zagoruyko2016wide}. From top to bottom, the feature map respectively come from the first layer to the fourth layer. Each feature map is labelled with its index in the layer and the corresponding AFS loss. From left to right, the localization performance of the feature maps gradually decreases. Our visualization intuitively demonstrates the localization bias caused by pre-training, indicating that not all feature maps contribute equally to anomaly detection and localization, as well as emphasizing the effectiveness of AFS.}
  \label{fig:figs7}
\end{figure*}

We conduct a comprehensive visual analysis of RealNet on the four datasets. \cref{fig:figs3} shows the qualitative results of RealNet in anomaly localization, showcasing its outstanding performance in pixel-level anomaly localization. Figs. \ref{fig:figs4} and \ref{fig:figs5} display the anomaly images and normal images generated by SDAS, respectively. \cref{fig:figs6} illustrates images synthesized using SIA with localized anomalous regions. \cref{fig:figs7} provides an intuitive explanation of pre-training bias, indicating that not all feature maps contribute equally to anomaly detection and localization, which validates the efficacy of AFS.

\section{Limitations}
\label{sec: supE}

In some categories with more texture anomalies, such as the texture categories in MVTec-AD dataset \cite{bergmann2019mvtec}, SIA's performance may slightly underperform when compared to DTD \cite{cimpoi2014describing}. Given that DTD dataset \cite{cimpoi2014describing} includes a diverse range of real-world texture images, it effectively simulates common anomaly types in the textural category, such as color, oil, and glue. Nonetheless, SIA excels in the majority of scenarios, outperforming DTD \cite{cimpoi2014describing} and offering superior capability in synthesizing anomalies in images with intricate structures.

Compared to anomaly synthesis methods based on data augmentation \cite{li2021cutpaste,schluter2022natural} or external data \cite{zavrtanik2021draem}, SDAS increases additional offline training time. For instance, we generate 10,000 anomaly images at a resolution of $256 \times 256$ for each category, and it will take 6 hours using a single NVIDIA GeForce RTX 3090. However, it is pivotal to clarify that RealNet omits SDAS without any additional computational cost during inference and real-world applications. Therefore, we believe that the slight increase in training time to enhance performance is necessary and worthwhile.

In order to achieve higher computational efficiency, we do not upsample multi-scale features. Instead, we employ multiple reconstruction networks for feature reconstruction, which reduce the resolution of anomaly detection. The lower feature reconstruction resolution may introduce minor deviations in localizing small anomalous areas, leading to a decrease in PRO. However, we found that increasing the resolution of anomaly detection by reducing the number of reconstruction networks can improve PRO. For instance, architecture \textbf{C} in \cref{fig:figs2} achieved a higher PRO score of 94.71\%. Furthermore, increasing the resolution of images can also lead to an improvement in PRO, as detailed in \cref{tab:tables4}.

\clearpage

\end{document}

%% file: preamble.tex
\usepackage[dvipsnames]{xcolor}

%% file: main.bbl
\begin{thebibliography}{55}
\providecommand{\natexlab}[1]{#1}
\providecommand{\url}[1]{\texttt{#1}}
\expandafter\ifx\csname urlstyle\endcsname\relax
  \providecommand{\doi}[1]{doi: #1}\else
  \providecommand{\doi}{doi: \begingroup \urlstyle{rm}\Url}\fi

\bibitem[Ak{\c{c}}ay et~al.(2019)Ak{\c{c}}ay, Atapour-Abarghouei, and
  Breckon]{akccay2019skip}
Samet Ak{\c{c}}ay, Amir Atapour-Abarghouei, and Toby~P Breckon.
\newblock Skip-ganomaly: Skip connected and adversarially trained
  encoder-decoder anomaly detection.
\newblock In \emph{2019 International Joint Conference on Neural Networks
  (IJCNN)}, pages 1--8. IEEE, 2019.

\bibitem[Baur et~al.(2019)Baur, Wiestler, Albarqouni, and Navab]{baur2019deep}
Christoph Baur, Benedikt Wiestler, Shadi Albarqouni, and Nassir Navab.
\newblock Deep autoencoding models for unsupervised anomaly segmentation in
  brain mr images.
\newblock In \emph{Brainlesion: Glioma, Multiple Sclerosis, Stroke and
  Traumatic Brain Injuries: 4th International Workshop, BrainLes 2018, Held in
  Conjunction with MICCAI 2018, Granada, Spain, September 16, 2018, Revised
  Selected Papers, Part I 4}, pages 161--169. Springer, 2019.

\bibitem[Bergmann et~al.(2019)Bergmann, Fauser, Sattlegger, and
  Steger]{bergmann2019mvtec}
Paul Bergmann, Michael Fauser, David Sattlegger, and Carsten Steger.
\newblock Mvtec-ad: A comprehensive real-world dataset for unsupervised anomaly
  detection.
\newblock In \emph{Proceedings of the IEEE/CVF conference on computer vision
  and pattern recognition}, pages 9592--9600, 2019.

\bibitem[Bergmann et~al.(2020)Bergmann, Fauser, Sattlegger, and
  Steger]{bergmann2020uninformed}
Paul Bergmann, Michael Fauser, David Sattlegger, and Carsten Steger.
\newblock Uninformed students: Student-teacher anomaly detection with
  discriminative latent embeddings.
\newblock In \emph{Proceedings of the IEEE/CVF conference on computer vision
  and pattern recognition}, pages 4183--4192, 2020.

\bibitem[Cimpoi et~al.(2014)Cimpoi, Maji, Kokkinos, Mohamed, and
  Vedaldi]{cimpoi2014describing}
Mircea Cimpoi, Subhransu Maji, Iasonas Kokkinos, Sammy Mohamed, and Andrea
  Vedaldi.
\newblock Describing textures in the wild.
\newblock In \emph{Proceedings of the IEEE conference on computer vision and
  pattern recognition}, pages 3606--3613, 2014.

\bibitem[Cohen and Hoshen(2020)]{cohen2020sub}
Niv Cohen and Yedid Hoshen.
\newblock Sub-image anomaly detection with deep pyramid correspondences.
\newblock \emph{arXiv preprint arXiv:2005.02357}, 2020.

\bibitem[Defard et~al.(2021)Defard, Setkov, Loesch, and
  Audigier]{defard2021padim}
Thomas Defard, Aleksandr Setkov, Angelique Loesch, and Romaric Audigier.
\newblock Padim: a patch distribution modeling framework for anomaly detection
  and localization.
\newblock In \emph{Pattern Recognition. ICPR International Workshops and
  Challenges: Virtual Event, January 10--15, 2021, Proceedings, Part IV}, pages
  475--489. Springer, 2021.

\bibitem[Deng and Li(2022)]{deng2022anomaly}
Hanqiu Deng and Xingyu Li.
\newblock Anomaly detection via reverse distillation from one-class embedding.
\newblock In \emph{Proceedings of the IEEE/CVF Conference on Computer Vision
  and Pattern Recognition}, pages 9737--9746, 2022.

\bibitem[Deng et~al.(2009)Deng, Dong, Socher, Li, Li, and
  Fei-Fei]{deng2009imagenet}
Jia Deng, Wei Dong, Richard Socher, Li-Jia Li, Kai Li, and Li Fei-Fei.
\newblock Imagenet: A large-scale hierarchical image database.
\newblock In \emph{2009 IEEE conference on computer vision and pattern
  recognition}, pages 248--255. Ieee, 2009.

\bibitem[Dhariwal and Nichol(2021)]{dhariwal2021diffusion}
Prafulla Dhariwal and Alexander Nichol.
\newblock Diffusion models beat gans on image synthesis.
\newblock \emph{Advances in Neural Information Processing Systems},
  34:\penalty0 8780--8794, 2021.

\bibitem[Duan et~al.(2023)Duan, Hong, Niu, and Zhang]{duan2023few}
Yuxuan Duan, Yan Hong, Li Niu, and Liqing Zhang.
\newblock Few-shot defect image generation via defect-aware feature
  manipulation.
\newblock In \emph{Proceedings of the AAAI Conference on Artificial
  Intelligence}, pages 571--578, 2023.

\bibitem[Gudovskiy et~al.(2022)Gudovskiy, Ishizaka, and
  Kozuka]{gudovskiy2022cflow}
Denis Gudovskiy, Shun Ishizaka, and Kazuki Kozuka.
\newblock Cflow-ad: Real-time unsupervised anomaly detection with localization
  via conditional normalizing flows.
\newblock In \emph{Proceedings of the IEEE/CVF Winter Conference on
  Applications of Computer Vision}, pages 98--107, 2022.

\bibitem[He et~al.(2016)He, Zhang, Ren, and Sun]{he2016deep}
Kaiming He, Xiangyu Zhang, Shaoqing Ren, and Jian Sun.
\newblock Deep residual learning for image recognition.
\newblock In \emph{Proceedings of the IEEE conference on computer vision and
  pattern recognition}, pages 770--778, 2016.

\bibitem[Heckler et~al.(2023)Heckler, K{\"o}nig, and
  Bergmann]{heckler2023exploring}
Lars Heckler, Rebecca K{\"o}nig, and Paul Bergmann.
\newblock Exploring the importance of pretrained feature extractors for
  unsupervised anomaly detection and localization.
\newblock In \emph{Proceedings of the IEEE/CVF Conference on Computer Vision
  and Pattern Recognition}, pages 2916--2925, 2023.

\bibitem[Heusel et~al.(2017)Heusel, Ramsauer, Unterthiner, Nessler, and
  Hochreiter]{heusel2017gans}
Martin Heusel, Hubert Ramsauer, Thomas Unterthiner, Bernhard Nessler, and Sepp
  Hochreiter.
\newblock Gans trained by a two time-scale update rule converge to a local nash
  equilibrium.
\newblock \emph{Advances in neural information processing systems}, 30, 2017.

\bibitem[Ho et~al.(2020)Ho, Jain, and Abbeel]{ho2020denoising}
Jonathan Ho, Ajay Jain, and Pieter Abbeel.
\newblock Denoising diffusion probabilistic models.
\newblock \emph{Advances in Neural Information Processing Systems},
  33:\penalty0 6840--6851, 2020.

\bibitem[Ioffe and Szegedy(2015)]{ioffe2015batch}
Sergey Ioffe and Christian Szegedy.
\newblock Batch normalization: Accelerating deep network training by reducing
  internal covariate shift.
\newblock In \emph{International conference on machine learning}, pages
  448--456. pmlr, 2015.

\bibitem[Jezek et~al.(2021)Jezek, Jonak, Burget, Dvorak, and
  Skotak]{jezek2021deep}
Stepan Jezek, Martin Jonak, Radim Burget, Pavel Dvorak, and Milos Skotak.
\newblock Deep learning-based defect detection of metal parts: evaluating
  current methods in complex conditions.
\newblock In \emph{2021 13th International Congress on Ultra Modern
  Telecommunications and Control Systems and Workshops (ICUMT)}, pages 66--71.
  IEEE, 2021.

\bibitem[Karras et~al.(2020)Karras, Laine, Aittala, Hellsten, Lehtinen, and
  Aila]{karras2020analyzing}
Tero Karras, Samuli Laine, Miika Aittala, Janne Hellsten, Jaakko Lehtinen, and
  Timo Aila.
\newblock Analyzing and improving the image quality of stylegan.
\newblock In \emph{Proceedings of the IEEE/CVF conference on computer vision
  and pattern recognition}, pages 8110--8119, 2020.

\bibitem[Li et~al.(2021)Li, Sohn, Yoon, and Pfister]{li2021cutpaste}
Chun-Liang Li, Kihyuk Sohn, Jinsung Yoon, and Tomas Pfister.
\newblock Cutpaste: Self-supervised learning for anomaly detection and
  localization.
\newblock In \emph{Proceedings of the IEEE/CVF Conference on Computer Vision
  and Pattern Recognition}, pages 9664--9674, 2021.

\bibitem[Liu et~al.(2023)Liu, Zhou, Xu, and Wang]{liu2023simplenet}
Zhikang Liu, Yiming Zhou, Yuansheng Xu, and Zilei Wang.
\newblock Simplenet: A simple network for image anomaly detection and
  localization.
\newblock In \emph{Proceedings of the IEEE/CVF Conference on Computer Vision
  and Pattern Recognition}, pages 20402--20411, 2023.

\bibitem[Liznerski et~al.(2021)Liznerski, Ruff, Vandermeulen, Franks, Kloft,
  and Muller]{liznerski2021explainable}
Philipp Liznerski, Lukas Ruff, Robert~A. Vandermeulen, Billy~Joe Franks, Marius
  Kloft, and Klaus~Robert Muller.
\newblock Explainable deep one-class classification.
\newblock In \emph{International Conference on Learning Representations}, 2021.

\bibitem[Lu et~al.(2023)Lu, Yao, Fu, and Jia]{Lu2023ICCV}
Fanbin Lu, Xufeng Yao, Chi-Wing Fu, and Jiaya Jia.
\newblock Removing anomalies as noises for industrial defect localization.
\newblock In \emph{Proceedings of the IEEE/CVF International Conference on
  Computer Vision (ICCV)}, pages 16166--16175, 2023.

\bibitem[Mishra et~al.(2021)Mishra, Verk, Fornasier, Piciarelli, and
  Foresti]{mishra2021vt}
Pankaj Mishra, Riccardo Verk, Daniele Fornasier, Claudio Piciarelli, and
  Gian~Luca Foresti.
\newblock Vt-adl: A vision transformer network for image anomaly detection and
  localization.
\newblock In \emph{2021 IEEE 30th International Symposium on Industrial
  Electronics (ISIE)}, pages 01--06. IEEE, 2021.

\bibitem[Nichol and Dhariwal(2021)]{nichol2021improved}
Alexander~Quinn Nichol and Prafulla Dhariwal.
\newblock Improved denoising diffusion probabilistic models.
\newblock In \emph{International Conference on Machine Learning}, pages
  8162--8171. PMLR, 2021.

\bibitem[P{\'e}rez et~al.(2003)P{\'e}rez, Gangnet, and Blake]{perez2003poisson}
Patrick P{\'e}rez, Michel Gangnet, and Andrew Blake.
\newblock Poisson image editing.
\newblock In \emph{ACM SIGGRAPH 2003 Papers}, pages 313--318. 2003.

\bibitem[Perlin(1985)]{perlin1985image}
Ken Perlin.
\newblock An image synthesizer.
\newblock \emph{ACM Siggraph Computer Graphics}, 19\penalty0 (3):\penalty0
  287--296, 1985.

\bibitem[Pirnay and Chai(2022)]{pirnay2022inpainting}
Jonathan Pirnay and Keng Chai.
\newblock Inpainting transformer for anomaly detection.
\newblock In \emph{Image Analysis and Processing--ICIAP 2022: 21st
  International Conference, Lecce, Italy, May 23--27, 2022, Proceedings, Part
  II}, pages 394--406. Springer, 2022.

\bibitem[Ristea et~al.(2022)Ristea, Madan, Ionescu, Nasrollahi, Khan, Moeslund,
  and Shah]{ristea2022self}
Nicolae-C{\u{a}}t{\u{a}}lin Ristea, Neelu Madan, Radu~Tudor Ionescu, Kamal
  Nasrollahi, Fahad~Shahbaz Khan, Thomas~B Moeslund, and Mubarak Shah.
\newblock Self-supervised predictive convolutional attentive block for anomaly
  detection.
\newblock In \emph{Proceedings of the IEEE/CVF Conference on Computer Vision
  and Pattern Recognition}, pages 13576--13586, 2022.

\bibitem[Roth et~al.(2022)Roth, Pemula, Zepeda, Sch{\"o}lkopf, Brox, and
  Gehler]{roth2022towards}
Karsten Roth, Latha Pemula, Joaquin Zepeda, Bernhard Sch{\"o}lkopf, Thomas
  Brox, and Peter Gehler.
\newblock Towards total recall in industrial anomaly detection.
\newblock In \emph{Proceedings of the IEEE/CVF Conference on Computer Vision
  and Pattern Recognition}, pages 14318--14328, 2022.

\bibitem[Schlegl et~al.(2017)Schlegl, Seeb{\"o}ck, Waldstein, Schmidt-Erfurth,
  and Langs]{schlegl2017unsupervised}
Thomas Schlegl, Philipp Seeb{\"o}ck, Sebastian~M Waldstein, Ursula
  Schmidt-Erfurth, and Georg Langs.
\newblock Unsupervised anomaly detection with generative adversarial networks
  to guide marker discovery.
\newblock In \emph{Information Processing in Medical Imaging: 25th
  International Conference, IPMI 2017, Boone, NC, USA, June 25-30, 2017,
  Proceedings}, pages 146--157. Springer, 2017.

\bibitem[Schl{\"u}ter et~al.(2022)Schl{\"u}ter, Tan, Hou, and
  Kainz]{schluter2022natural}
Hannah~M Schl{\"u}ter, Jeremy Tan, Benjamin Hou, and Bernhard Kainz.
\newblock Natural synthetic anomalies for self-supervised anomaly detection and
  localization.
\newblock In \emph{Computer Vision--ECCV 2022: 17th European Conference, Tel
  Aviv, Israel, October 23--27, 2022, Proceedings, Part XXXI}, pages 474--489.
  Springer, 2022.

\bibitem[Shi et~al.(2021)Shi, Yang, and Qi]{yang2020dfr}
Yong Shi, Jie Yang, and Zhiquan Qi.
\newblock Unsupervised anomaly segmentation via deep feature reconstruction.
\newblock \emph{Neurocomputing}, 424:\penalty0 9--22, 2021.

\bibitem[Song et~al.(2021)Song, Meng, and Ermon]{song2020denoising}
Jiaming Song, Chenlin Meng, and Stefano Ermon.
\newblock Denoising diffusion implicit models.
\newblock In \emph{International Conference on Learning Representations}, 2021.

\bibitem[Tan and Le(2019)]{tan2019efficientnet}
Mingxing Tan and Quoc Le.
\newblock Efficientnet: Rethinking model scaling for convolutional neural
  networks.
\newblock In \emph{International conference on machine learning}, pages
  6105--6114. PMLR, 2019.

\bibitem[Tang et~al.(2020)Tang, Kuo, Lan, Ding, Hsu, and
  Young]{tang2020anomaly}
Ta-Wei Tang, Wei-Han Kuo, Jauh-Hsiang Lan, Chien-Fang Ding, Hakiem Hsu, and
  Hong-Tsu Young.
\newblock Anomaly detection neural network with dual auto-encoders gan and its
  industrial inspection applications.
\newblock \emph{Sensors}, 20\penalty0 (12):\penalty0 3336, 2020.

\bibitem[Tao et~al.(2022)Tao, Zhang, Ma, Hou, Lu, and
  Adak]{tao2022unsupervised}
Xian Tao, Dapeng Zhang, Wenzhi Ma, Zhanxin Hou, ZhenFeng Lu, and Chandranath
  Adak.
\newblock Unsupervised anomaly detection for surface defects with dual-siamese
  network.
\newblock \emph{IEEE Transactions on Industrial Informatics}, 18\penalty0
  (11):\penalty0 7707--7717, 2022.

\bibitem[Tien et~al.(2023)Tien, Nguyen, Tran, Huy, Duong, Nguyen, and
  Truong]{tien2023revisiting}
Tran~Dinh Tien, Anh~Tuan Nguyen, Nguyen~Hoang Tran, Ta~Duc Huy, Soan Duong,
  Chanh D~Tr Nguyen, and Steven~QH Truong.
\newblock Revisiting reverse distillation for anomaly detection.
\newblock In \emph{Proceedings of the IEEE/CVF Conference on Computer Vision
  and Pattern Recognition}, pages 24511--24520, 2023.

\bibitem[Wyatt et~al.(2022)Wyatt, Leach, Schmon, and
  Willcocks]{wyatt2022anoddpm}
Julian Wyatt, Adam Leach, Sebastian~M Schmon, and Chris~G Willcocks.
\newblock Anoddpm: Anomaly detection with denoising diffusion probabilistic
  models using simplex noise.
\newblock In \emph{Proceedings of the IEEE/CVF Conference on Computer Vision
  and Pattern Recognition}, pages 650--656, 2022.

\bibitem[Xi et~al.()Xi, Liu, Wang, Nie, Kai, Liu, Wang, and Zheng]{xisoftpatch}
Jiang Xi, Jianlin Liu, Jinbao Wang, Qiang Nie, WU Kai, Yong Liu, Chengjie Wang,
  and Feng Zheng.
\newblock Softpatch: Unsupervised anomaly detection with noisy data.
\newblock In \emph{Advances in Neural Information Processing Systems}.

\bibitem[Yang et~al.(2023)Yang, Wu, and Feng]{yang2023memseg}
Minghui Yang, Peng Wu, and Hui Feng.
\newblock Memseg: A semi-supervised method for image surface defect detection
  using differences and commonalities.
\newblock \emph{Engineering Applications of Artificial Intelligence},
  119:\penalty0 105835, 2023.

\bibitem[Yao et~al.(2023)Yao, Li, Zhang, Sun, and Zhang]{yao2023explicit}
Xincheng Yao, Ruoqi Li, Jing Zhang, Jun Sun, and Chongyang Zhang.
\newblock Explicit boundary guided semi-push-pull contrastive learning for
  supervised anomaly detection.
\newblock In \emph{Proceedings of the IEEE/CVF Conference on Computer Vision
  and Pattern Recognition}, pages 24490--24499, 2023.

\bibitem[Yi and Yoon(2020)]{yi2020patch}
Jihun Yi and Sungroh Yoon.
\newblock Patch svdd: Patch-level svdd for anomaly detection and segmentation.
\newblock In \emph{Proceedings of the Asian Conference on Computer Vision},
  2020.

\bibitem[You et~al.(2022)You, Cui, Shen, Yang, Lu, Zheng, and
  Le]{you2022unified}
Zhiyuan You, Lei Cui, Yujun Shen, Kai Yang, Xin Lu, Yu Zheng, and Xinyi Le.
\newblock A unified model for multi-class anomaly detection.
\newblock In \emph{Advances in Neural Information Processing Systems}, 2022.

\bibitem[Youkachen et~al.(2019)Youkachen, Ruchanurucks, Phatrapomnant, and
  Kaneko]{youkachen2019defect}
Sanyapong Youkachen, Miti Ruchanurucks, Teera Phatrapomnant, and Hirohiko
  Kaneko.
\newblock Defect segmentation of hot-rolled steel strip surface by using
  convolutional auto-encoder and conventional image processing.
\newblock In \emph{2019 10th International Conference of Information and
  Communication Technology for Embedded Systems (IC-ICTES)}, pages 1--5. IEEE,
  2019.

\bibitem[Yu et~al.(2021)Yu, Zheng, Wang, Li, Wu, Zhao, and Wu]{yu2021fastflow}
Jiawei Yu, Ye Zheng, Xiang Wang, Wei Li, Yushuang Wu, Rui Zhao, and Liwei Wu.
\newblock Fastflow: Unsupervised anomaly detection and localization via 2d
  normalizing flows.
\newblock \emph{arXiv preprint arXiv:2111.07677}, 2021.

\bibitem[Zagoruyko and Komodakis(2016)]{zagoruyko2016wide}
Sergey Zagoruyko and Nikos Komodakis.
\newblock Wide residual networks.
\newblock In \emph{Proceedings of the British Machine Vision Conference
  (BMVC)}, pages 87.1--87.12. BMVA Press, 2016.

\bibitem[Zavrtanik et~al.(2021)Zavrtanik, Kristan, and
  Sko{\v{c}}aj]{zavrtanik2021draem}
Vitjan Zavrtanik, Matej Kristan, and Danijel Sko{\v{c}}aj.
\newblock Draem: A discriminatively trained reconstruction embedding for
  surface anomaly detection.
\newblock In \emph{Proceedings of the IEEE/CVF International Conference on
  Computer Vision}, pages 8330--8339, 2021.

\bibitem[Zavrtanik et~al.(2022)Zavrtanik, Kristan, and
  Sko{\v{c}}aj]{zavrtanik2022dsr}
Vitjan Zavrtanik, Matej Kristan, and Danijel Sko{\v{c}}aj.
\newblock Dsr: A dual subspace re-projection network for surface anomaly
  detection.
\newblock In \emph{Computer Vision--ECCV 2022: 17th European Conference, Tel
  Aviv, Israel, October 23--27, 2022, Proceedings, Part XXXI}, pages 539--554.
  Springer, 2022.

\bibitem[Zhang et~al.(2023{\natexlab{a}})Zhang, Wu, Wang, Chen, and
  Jiang]{zhang2023prototypical}
Hui Zhang, Zuxuan Wu, Zheng Wang, Zhineng Chen, and Yu-Gang Jiang.
\newblock Prototypical residual networks for anomaly detection and
  localization.
\newblock In \emph{Proceedings of the IEEE/CVF Conference on Computer Vision
  and Pattern Recognition}, pages 16281--16291, 2023{\natexlab{a}}.

\bibitem[Zhang et~al.(2018)Zhang, Isola, Efros, Shechtman, and
  Wang]{zhang2018unreasonable}
Richard Zhang, Phillip Isola, Alexei~A Efros, Eli Shechtman, and Oliver Wang.
\newblock The unreasonable effectiveness of deep features as a perceptual
  metric.
\newblock In \emph{Proceedings of the IEEE conference on computer vision and
  pattern recognition}, pages 586--595, 2018.

\bibitem[Zhang et~al.(2023{\natexlab{b}})Zhang, Li, Li, Dai, Jiang, and
  Xia]{Zhang2023ICCV}
Xinyi Zhang, Naiqi Li, Jiawei Li, Tao Dai, Yong Jiang, and Shu-Tao Xia.
\newblock Unsupervised surface anomaly detection with diffusion probabilistic
  model.
\newblock In \emph{Proceedings of the IEEE/CVF International Conference on
  Computer Vision (ICCV)}, pages 6782--6791, 2023{\natexlab{b}}.

\bibitem[Zhang et~al.(2023{\natexlab{c}})Zhang, Li, Li, Huang, Shan, and
  Chen]{zhang2023destseg}
Xuan Zhang, Shiyu Li, Xi Li, Ping Huang, Jiulong Shan, and Ting Chen.
\newblock Destseg: Segmentation guided denoising student-teacher for anomaly
  detection.
\newblock In \emph{Proceedings of the IEEE/CVF Conference on Computer Vision
  and Pattern Recognition}, pages 3914--3923, 2023{\natexlab{c}}.

\bibitem[Zhao(2023)]{zhao2023omnial}
Ying Zhao.
\newblock Omnial: A unified cnn framework for unsupervised anomaly
  localization.
\newblock In \emph{Proceedings of the IEEE/CVF Conference on Computer Vision
  and Pattern Recognition}, pages 3924--3933, 2023.

\bibitem[Zou et~al.(2022)Zou, Jeong, Pemula, Zhang, and Dabeer]{zou2022spot}
Yang Zou, Jongheon Jeong, Latha Pemula, Dongqing Zhang, and Onkar Dabeer.
\newblock Spot-the-difference self-supervised pre-training for anomaly
  detection and segmentation.
\newblock In \emph{Computer Vision--ECCV 2022: 17th European Conference, Tel
  Aviv, Israel, October 23--27, 2022, Proceedings, Part XXX}, pages 392--408.
  Springer, 2022.

\end{thebibliography}
